\documentclass[journal]{IEEEtran}

\usepackage[linesnumbered,ruled,vlined]{algorithm2e}
\usepackage{multirow,booktabs,color,soul}
\usepackage{booktabs}
\usepackage{multirow}
\usepackage{subfigure}
\usepackage{graphicx}
\usepackage{amsmath}
\usepackage{amssymb}
\usepackage{textcomp}

\ifCLASSINFOpdf
\else
\fi

\begin{document}

\title{A Two-stage Framework and Reinforcement Learning-based Optimization Algorithms for Complex Scheduling Problems}

\author{Yongming He, Guohua Wu, Yingwu Chen, and Witold Pedrycz,~\IEEEmembership{Fellow,~IEEE}

\thanks{Yongming He is with the College of Systems Engineering, National University of Defense Technology, Changsha 410073, P. R. China. E-mail: heyongming10@hotmail.com}

\thanks{Guohua Wu (\textit{the corresponding author}) is with the School of Traffic and Transportation Engineering, Central South University, Changsha 410075, China. E-mail: guohuawu@csu.edu.cn}

\thanks{Yingwu Chen is with the College of Systems Engineering, National University of Defense Technology, Changsha 410073, P. R. China. E-mail: ywchen@nudt.edu.cn}

\thanks{Witold Pedrycz is with the Department of Electrical and Computer Engineering, University of Alberta, Edmonton, AB T6G 2V4, Canada, and also with the Systems Research Institute, Polish Academy of Sciences, Warsaw 01447, Poland. E-mail: wpedrycz@ualberta.ca}
}


\markboth{IEEE Transactions on Cybernetics}%
{Yongming \MakeLowercase{\\textit{et al.}}: Bare Demo of IEEEtran.cls for IEEE Journals}

\maketitle

\begin{abstract}

There hardly exists a general solver that is efficient for scheduling problems due to their diversity and complexity. In this study, we develop a two-stage framework, in which reinforcement learning (RL) and traditional operations research (OR) algorithms are combined together to efficiently deal with complex scheduling problems. The scheduling problem is solved in two stages, including a finite Markov decision process (MDP) and a mixed-integer programming process, respectively. This offers a novel and general paradigm that combines RL with OR approaches to solving scheduling problems, which leverages the respective strengths of RL and OR: The MDP narrows down the search space of the original problem through an RL method, while the mixed-integer programming process is settled by an OR algorithm. These two stages are performed iteratively and interactively until the termination criterion has been met. Under this idea, two implementation versions of the combination methods of RL and OR are put forward. The agile Earth observation satellite scheduling problem is selected as an example to demonstrate the effectiveness of the proposed scheduling framework and methods. The convergence and generalization capability of the methods are verified by the performance of training scenarios, while the efficiency and accuracy are tested in 50 untrained scenarios. The results show that the proposed algorithms could stably and efficiently obtain satisfactory scheduling schemes for agile Earth observation satellite scheduling problems. In addition, it can be found that RL-based optimization algorithms have stronger scalability than non-learning algorithms. This work reveals the advantage of combining reinforcement learning methods with heuristic methods or mathematical programming methods for solving complex combinatorial optimization problems.

\end{abstract}

\begin{IEEEkeywords}

Scheduling, operations research, reinforcement learning, two-stage optimization.

\end{IEEEkeywords}

\IEEEpeerreviewmaketitle

\section{Introduction}\label{Section1}

\IEEEPARstart{S}{cheduling} problems are widespread in production, manufacturing, transportation, and logistics, which are typically formulated as combinatorial optimization models. This type of model generally contains an objective function and a set of constraints, such as job-shop scheduling problem (JSP)\cite{gambardella1995ant}, traveling salesman problem with time window (TSPTW)\cite{kona2015review}, and Earth observation satellite scheduling problems\cite{du2019data}. Mathematical optimization models such as mixed-integer linear programming and nonlinear integer programming are popular for depicting concrete scheduling problems\cite{ji2018hybrid}, but it is hard to find a relatively general and efficient algorithm to solve all of them, due to the significant difference in optimization objectives and the constraints of miscellaneous scheduling problems.

Tremendous number of scheduling problems in applications are proved to be NP-hard. In this study, a type of complex and representative problem is particularly concerned, which contains two levels of decision variables, i.e., 1) Which resource these tasks should be executed? 2) When should each candidate task execute? 

For a long time, mathematical programming, heuristic and metaheuristic algorithms were the main tools to tackle the above scheduling problems\cite{michael2018scheduling}. However, the bottlenecks of these algorithms become prominent with the scale and the complexity of the problem increase: either the computing overhead is unacceptable, or the solution is unsatisfactory or unstable\cite{du2019data,lu2019learning}. Furthermore, the construction of rules or parameters in these algorithms is closely related to the characteristics of the problem, which leads to the fact that these algorithms are not versatile. Thus, a “general-purpose” and “far-sighted” algorithm with short decision time is needed. 

The scheduling problem considered in this study actually can be decomposed into following three sub-problems\cite{wolfe2000three}: 1) Assignment problem, for determining the possible tasks for each resource, i.e. assign tasks to different resources; 2) Sequencing problem, for determining the order of these tasks at each resource; 3) Timing problem, for determining the execution start and end time of scheduled tasks at each resource. These three sub-problems are mutually correlated, and the original scheduling problem could be solved by addressing them reasonably. It is a promising way to divide the original scheduling problems into two or multiple stages and then solve them separately, which could reduce the overall complexity and solve the scheduling problem more efficiently\cite{wang2019joint}. In particular, a two-stage framework is built for dealing with the considered scheduling problems.

The proposed two-stage framework consists of a prior stage and a rear stage: the prior stage for assignment problem, while the rear stage for sequencing problem and timing problem. It is desirable and natural to model the prior stage as a finite Markov decision process (MDP) and then solving it by reinforcement learning (RL): 1) The model can be well trained in advance by RL and achieve fast and accurate decisions in unknown scenarios; 2) It learns scheduling policy by trading off between immediate rewards and delayed value, and therefore RL makes decisions by weighing the estimated returns of each part of the system; 3) It is not required to manually design the decision-making rules for each instance. By contrast, the sub-problem in the rear stage can be formulated as a mixed-integer programming model, and then be solved by traditional operations research (OR) techniques.

For examining the effectiveness of our approach, deep Q-learning and an OR algorithm (i.e., a constructive heuristic or dynamic programming) are integrated the two-stage framework to solve the considered complex scheduling problem. The agile satellite scheduling problem is taken as an example to demonstrate the potential in applying the two-stage framework and the RL-based optimization algorithms. Experiments show that the two-stage scheduling framework is efficient in dealing with complex scheduling problems in the real-world.

The major contributions of this paper  are:

\begin{itemize}
	
	\item A novel two-stage scheduling framework is proposed for decomposing a type of complex scheduling problem into two parts,  which are formulated by an MDP model and a mixed-integer programming model, respectively.
	
	\item An RL-based optimization mechanism is proposed under the two-stage framework. Specifically, deep Q-learning and OR algorithms (i.e., a constructive heuristic algorithm or a dynamic programming (DP) algorithm) are employed to solve the MDP model and mixed-integer programming model, respectively.
	
	\item The effectiveness of the proposed two-stage framework and RL-based optimization algorithms are verified by applying them to agile Earth observation satellite scheduling problem. In addition, some general conclusions about the model, mechanism, and algorithms are summarized.
	
\end{itemize}

The rest of the paper is structured as follows: Section~\ref{Section2} reviews existing studies on two-stage optimization models, scheduling algorithms, as well as reinforcement learning applications in combinatorial optimization problems. A general two-stage framework for scheduling problems is proposed in Section~\ref{Section3}. RL-based optimization methods are detailed in Section~\ref{Section4}. Section~\ref{Section5} discusses the convergence and the generalization capabilities of proposed methods and then compares the proposed methods with other algorithms.

\section{Literature review}\label{Section2}

As an important branch in the field of operations research, scheduling problems have attracted much attention from the scholars for a long time. After decades of research, scholars have been working to find more accurate and faster approaches to solve them, and some significant scheduling problems have been formed: resource-constrained project scheduling problem\cite{hartmann2010survey}, vehicle routing problem with time windows (VRPTW)\cite{dixit2019vehicle}, job-shop scheduling problem (JSP)\cite{jones2001survey}, and so on. Several ways for formulating these problems were proposed, such as the mathematical programming models\cite{bracken1973mathematical, bertsimas1997introduction} and the constraint satisfaction model\cite{verfaillie1996russian}. With the increasing real-life considerations in scheduling problems, the above models usually are hard to describe and simplify as constraints vary over different situations\cite{lachhwani2018bi, turner2017distributed}. Plilippe Baptiste et al.\cite{baptiste2006constraint} tried to summarize the constraints of scheduling problems into several categories, but many constraints in practical scheduling processes are still difficult to handle. 

Because of the difficulty of many real-world scheduling problems, two-stage scheduling methods came into being. Two-stage decision-making architecture is common-used in practical scheduling projects, e.g., staff scheduling\cite{tsai2009two, parisio2015two}, satellite task scheduling\cite{deng2017two}, issues in the transportation field\cite{wu2017two}, and extensive optimization problems in industry\cite{qiu2018bi, dashti2016weekly}. However, solutions to these works are usually application-oriented. Hence, designing a novel and generic two-stage framework is meaningful to study scheduling problems in a better way.

It is of great significance to  make satisfactory decisions with short computing time, especially when responding to scenario changes. Traditional algorithms for scheduling problems can be summarised into three categories: mathematical programming, heuristic algorithms, and metaheuristic algorithms. Mathematical programming such as branch-and-bound\cite{chu2017branch}, branch-and-price\cite{vaclavik2018accelerating}, and dynamic programming\cite{xiao2017genetic} guarantee the optimal solution under certain assumptions, but the time complexity of these algorithms is usually exponential. Therefore, they are usually time-consuming and space-consuming in dealing with large scale problems. Heuristic algorithms find a solution by constructive rules for decision making. These rules are usually needed to be designed by experts based on experience. Heuristic algorithms can obtain a feasible solution in a short time. However, they cannot guarantee optimality, and may perform unsatisfactory in some situations. Furthermore, the search strategies of both exact algorithms and heuristic algorithms are highly coupled to the features and conditions of the concrete model considered. Metaheuristics such as tabu search\cite{braysy2002tabu}, genetic algorithm\cite{tsai2009two}, adaptive large neighborhood search (ALNS)\cite{liu2017adaptive} and their variants spring up due to their competive performance in tackling complex scheduling problems. However, they are difficult to improve time efficiency and solution accuracy simultaneously, thereby the solutions may be unsatisfactory and unstable in affordable computation time.

It is no longer new about solving scheduling problems based on RL: Luca et.al.\cite{gambardella1995ant} proposed ant-Q algorithm for solving traveler salesman problem (TSP); Wei and Zhao\cite{wei2005reinforcement} used Q-learning to select composite rule (a machine-job pair) in job-shop problem (JSP); Khalil et al.\cite{khalil2017learning} suggested deep reinforcement learning (GNN-based deep Q-learning) to train the value function of each step of decision in the TSP problem;  Nazari et.al.\cite{nazari2018reinforcement} solved vehicle routing problem (VRP) by pointer networks; Li et.al.\cite{li2020deep} applied an improved actor-critic algorithm to train the model for multi-objective travelling salesman problem. These works show that RL has great potential in solving scheduling problems. However, the required number of iterations to reach satisfactory results rises sharply when RL is applied to intricate problems\cite{lu2019learning}, which may be unacceptable in many real-world applications. Decomposing the original problem and solving some sub-problems by OR algorithms can reduce the solution search space and improve learning speed of RL. 

\section{The two-stage scheduling framework}\label{Section3}

For the sake of uniform and clear description of the type of scheduling problems considered in this study, we first introduce and discuss its characteristics and then build a two-stage scheduling framework for solving the problem efficiently.

\subsection{Notation}

Table~\ref{Table:Notation} lists all necessary notation required to describe the scheduling problem, while algorithm-related notation will be illustrated where they occur.

\begin{table}[h]
	
	\caption{Notation} \label{Table:Notation}
	
	\begin{tabular}{cc}
		\toprule
		Notation & Interpretation \\
		\midrule
		$i$ & Task index\\	
		$j$ & Resource index\\	
		$n$ & The number of tasks\\	
		$m$ & The number of resources\\	
		$\boldsymbol{ TS }$ & The set of tasks\\	
		$\boldsymbol{ RS }$ & The set of resources\\	
		$C$ & The capacity of resources\\	
		$R _ i$ & Reward at time step $i$\\	
		$\left[ws_i^j,we_i^j\right]$ & Executable time window of task $i$ on resource $j$\\	
		$d_i^j$ & Execution duration of task $i$ on resource $j$\\	
		$p_i^j$ & Profit of task $i$ on resource $j$\\	
		$\Omega$ & Solution space of the original problem\\	
		$\Omega_1$ & Solution space of the problem in the prior stage\\	
		$\Omega_2$ & Solution space of the problem in the rear stage\\			
		$t$ & Time step in decision process\\	
		$S_t$ & State at time step $t$ in the MDP\\	
		$s_i^t$ & State of task $i$ at time step $t$ in the MDP\\	
		$R_t$ & Reward at time step $t$ in the MDP\\	
		$\boldsymbol{ A }$ & Action space of the MDP\\	
		$\boldsymbol{ A }\left(S_t\right)$ & Available action set under state $S_t$\\	
		$w_i^t$ & Remain executable time windows of task $i$ at time step $t$\\
		\bottomrule
	\end{tabular}
\end{table}

\subsection{Problem description}

\begin{figure}[h]
	\begin{center}
		\includegraphics[width=0.48\textwidth]{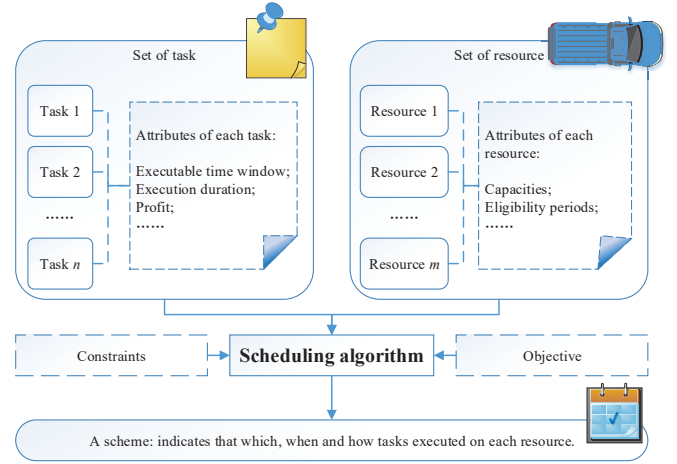}
	\end{center}
	\caption{A brief description of scheduling problems.} \label{Fig:ProblemDesc}
\end{figure}

The concerned scheduling problems can be defined uniformly as follows\cite{michael2018scheduling}: There are a certain number of resources, each with certain capacities and restrictions in their usages. There are a certain number of tasks, each of which could associated with several records of executable time windows (ETWs), execution durations on each resource, and the profit of being completed. The problem is to assign the resource for scheduled tasks and then determine their execution start and end time, to optimize the objective function and meet all constraints. A schematic diagram of the problem is shown in Fig~\ref{Fig:ProblemDesc}.

The input of a scheduling problem usually consists of two parts: resource information and task information. 

Resource information describes the set of resources. Each resource may have capacity restrictions in multiple dimensions, e.g., The total amount of cargo that can be loaded on each vehicle in the VRP problem is limited by both volume and weight. The set of resources information is formalized as follow:
\begin{equation}
\boldsymbol{ RS } = \left\{ RS _ { 1 } , RS _ { 2 } , \ldots , RS _ { m } \right\} \label{res}
\end{equation}
\begin{equation}
RS _ j = \left( C _ { 1 } , C _ { 2 } , \ldots \right)\label{resj}
\end{equation}

A task $TS _ i$ can be described by a set of triples: the ETW $[ws _ i ^ j, we _ i ^ j]$, required execution duration $d _ i ^ j$, and the profit $p _ i ^ j$. These attributes are related to the task $i$ and resource $j$. The set of tasks is formalized as follow:
\begin{equation}
TS _ { i } ^ { j } = \left\{ \begin{array} { c l } { \left(\left[w s_{i}^{j}, w e_{i}^{j}\right], d_{i}^{j}, p_{i}^{j}\right)} & { \text {, } i \text{ is available in } j } \\ { \emptyset } & {  \text {, otherwise }} \end{array} \right. \label{tsij}
\end{equation}
\begin{equation}
TS _ i=\cup_{j=1,2, \ldots, m} T S_{i}^{j}\label{tsi}
\end{equation}
\begin{equation}
\boldsymbol{TS}=\cup_{i=1,2, \ldots, n} T S_{i}\label{ts}
\end{equation}

In general, the execution start time and the service resource are the decision variables for each scheduled task in the considered problem. They are represented by $es _ i$ and $r _ i$, respectively.

$r _ i$ is a variable indicating the resource on which task $i$ is scheduled.
\begin{equation}
r _ { i } = \left\{ \begin{array} { l l } { -1 , } & { \text { task } i \text { is unscheduled }} \\ { j , } & { \text { task } i \text { is scheduled on resource } j } \end{array} \right. \label{ri}
\end{equation}

$es _ i$ and $ee _ i$ indicate the execution start and end time of task $i$, and the values of these two variables are $null$ if and only if $r _ i = -1$. Otherwise, we have
\begin{equation}
ee _ { i } = es _ {i} + d _ i ^ {r _ i}, r _ i \neq -1 \label{eei}
\end{equation}

If $es_i \neq null$ and $ee_i \neq null$, the relationship between $ws_i^j$, $we_i^j$, $es_i$ and $ee_i$ is:
\begin{equation}
w s_{i}^{j} \leq e s_{i}<e e_{i} \leq w e_{i}^{j}, \exists j \in \boldsymbol{ RS }\label{wsij}
\end{equation}

\subsection{Two-stage representation for scheduling problems}

Let us refer to the definition of two-stage optimization problem\cite{goerigk2020two, vicente1994bilevel, lu2016multilevel}. The two-stage scheduling model can be described as (\ref{Model:origin}): 
\begin{equation}
\begin{array} { l l } { \min _{\boldsymbol{r}, \boldsymbol{es}} } & { F(\boldsymbol{r}, \boldsymbol{es}) } \\ { \text{s.t.} } & {  G_{1}(\boldsymbol{r}) \leq \mathbf{0} } \\ {  } & {  G_{2}(\boldsymbol{r}, \boldsymbol{e s}) \leq \mathbf{0} } \\ {  } & { \boldsymbol{ r } = \left(r _ 1 , r _ 2 , \ldots , r _ n\right) \in \Omega _ 1 } \\ {  } & {  \boldsymbol{ es } = \left(es _ 1 , es _ 2 , \ldots , es _ n\right) \in \Omega _ 2 } \end{array}\label{Model:origin}
\end{equation}
where, $F(\boldsymbol{r}, \boldsymbol{es})$ is the objective function of the model; $G_{1}(\boldsymbol{r})$ and $G_{2}(\boldsymbol{r}, \boldsymbol{es})$ refers the constraints of  the prior stage and the rear stage; $\boldsymbol{ r }$ and $\boldsymbol{ es }$ are decision variables; $\Omega _ 1$ and $\Omega _ 2$ denote the research spaces of the prior stage and the rear stage, respectively. The solution space of the original problem $\Omega$ is calculated by the vector product of the solution space of two stages: $\Omega = \Omega_1 \times \Omega_2$. The feasible set of the rear stage could be reduced by fixing $\boldsymbol{ r }$.
\begin{equation}
\Omega_{2}\left(\boldsymbol{r}^{\prime}\right)=\left\{\boldsymbol{es} \mid G_{2}\left(\boldsymbol{r}^{\prime}, \boldsymbol{es}\right) \leq \boldsymbol{0}\right\}
\end{equation}
where, $\boldsymbol{r}^{\prime}$ is a fixed value of $\boldsymbol{ r }$.

Generally, the rear stage can be formed as an optimization problem individually and solved by some optimization methods, as the problem of the rear stage is described as follows.
\begin{equation}
\begin{array} { l l } { \min _{\boldsymbol{ es }\in\Omega_{2}(\boldsymbol{r}^{\prime} )} } & { F(\boldsymbol{{r}^{\prime}}, \boldsymbol{es}) } \\ { \text{s.t.} } &  {  G_{2}(\boldsymbol{{r}^{\prime}}, \boldsymbol{e s}) \leq \mathbf{0} } \end{array}\label{Model:rear}
\end{equation}

However, the prior stage is hard to deal with in this way. The objective of the prior stage is the same as the objective of the original problem, and the objective value is strongly coupling with the results on the rear stage. In this study, we model the problem in the prior stage as an MDP model.

\subsubsection{Action}

A solution of the prior stage $\boldsymbol{ r }$ can be transformed to $\mathcal{A}_t$, which is a set of actions before any time step $t$, i.e.
\begin{equation}
\mathcal{A}_t = \left\{a_0,a_1,\dots,a_t\right\},a_i\in\boldsymbol{ A }\left(S_i\right),i = 0,1,\ldots,t\label{action}
\end{equation}

Actions in this model contain ``Selecting a task for a resource'' and ``Termination''. ``Selecting a task for a resource'' is the basic action in this model. Before taking an action at a time step, constraints which are unrelated to decision variables are checked before the task is selected for reducing unnecessary calculations. These constraints may include but are not limited to:

\begin{itemize}
	\item Executable time window restriction;
	\item Consumption of each task and the capacity of the resource;
\end{itemize}

``Termination'' is an action that means the task selection within a certain planning cycle of this resource is completed. In the case of being without any available task at time step $t$, ``Termination'' is the only action. 

\subsubsection{State}

A state depicts the attributes of each task on a determined time point of decision-making, it forms as Eq.~(\ref{state}):
\begin{equation}
S _ { t } = \left\{ s _ { i } ^ { t } = \left( w _ { i } ^ { t } , d _ { i } ^ { t } , p _ { i } ^ { t } \right) | t = 0,1 , \ldots , n \right\} \label{state}
\end{equation}
where $w _ { i } ^ { t }$ is the set of remain ETWs of task $i$ at time step $t$; $d _ { i } ^ { t }$ represents the duration of task $i$ at time step $t$;  $p _ { i } ^ { t }$ stands for the profit of completing task $i$ at time step $t$.

The terminal states of scheduling problems are commonly expressed as Eq.~(\ref{terminalstate}), that is, there are no ETW remaining for all tasks.
\begin{equation}
w _ { i } ^ { t } = \emptyset , \forall i \in \left\{1,2,\ldots, n \right\} \label{terminalstate}
\end{equation}

\subsubsection{Reward}

Reward indicates the difference in the value of the objective function between a step and its previous step. It can be counted by Eq.~(\ref{reward})
\begin{equation}
R _ { t } = F\left(\mathcal{ A } _ t,\boldsymbol{ es }\right)-F\left(\mathcal{ A } _ {t - 1},\boldsymbol{ es }\right) \label{reward}
\end{equation}
where $F\left(\mathcal{ A } _ t,\boldsymbol{ es }\right)$ and $F\left(\mathcal{ A } _ {t - 1},\boldsymbol{ es }\right)$ are the values of objective function at time step $t$ and $t - 1$, respectively.

\subsubsection{Value function}

Value function refers to the expected long term profit under the condition of determined state and action, which is the criterion of taking actions. Due to the huge size of the state, tabular solution methods are hard to reflect the nature of the value function and support decisions efficiently. Artificial neural networks (ANNs) are popular and efficient  to represent the value function owing to their strong prediction and classification capabilities\cite{bengio2018machine, nguyen2018deep}, so we deem that ANN could be a promising tool for estimating the value function. 

The input of the network is the state of the model, while the output is the expected long term profit of each action. The value function is continuously updated according to a large number of records about actions, states, and rewards. 

\section{Reinforcement learning-based optimization algorithms}\label{Section4}

\begin{figure}[h]
	\begin{center}
		\includegraphics[width=0.48\textwidth]{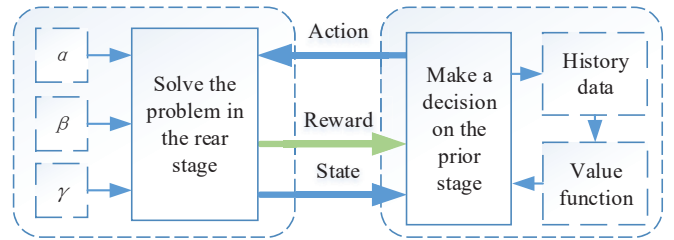}
	\end{center}
	\caption{The interactions between two stages of the process. $\alpha,\beta,$ and $\gamma$ draw the environment, objective, and constraints of the mixed-integer programming problem, respectively; Action, Reward, and State are defined in Section~\ref{Section3}.} \label{Fig:Interactions}
\end{figure}

We decompose the scheduling problem into two stages, i.e., a finite Markov decision process for assignment problem while a mixed-integer programming process for sequencing problem and timing problem. The interactions between these two processes are that mixed-integer programming process obtains a solution according to the results of finite MDP; while the finite MDP makes decisions considering the results of the mixed-integer programming. During training phase, the value function of the MDP is continually fitted based on records of decisions and results, for a better decision criterion. The relationship between the two stages is shown in Fig.~\ref{Fig:Interactions}.

We design the deep Q-learning (DQN) for the assignment problem in the prior stage, and an OR algorithm (a constructive heuristic (HADRT) or dynamic programming(DP)) for the sequencing and timing problem in the rear stage. The approaches with HADRT and DP are named DQN\_CH and DQN\_DP, respectively.

\subsection{DQN in the prior stage}

There is a wide range of state of scheduling problems. In the other hand, the initial states of a scheduling problem vary because of the changable attributes of tasks and resources, so it is necessary to get adequate knowledge in limited iterations and to make decisions under unknown states.

The training process is usually organized as the following steps: 

\textbf{Step 1}: Choose an action by ``Exploitation'' or ``Exploration'';

\textbf{Step 2}: If the action is not ``Termination'', get the reward and state by interactions between two stages;

\textbf{Step 3}: Update the value function based on actions, rewards, and states; 

\textbf{Step 4}: Repeat \textbf{Step 1} to \textbf{Step 3} until the termination state is reached. 

The policy for choosing actions includes two strategies, i.e., ``Exploitation'' and ``Exploration''. ``Exploitation'' signifies choosing the best action according to the predicted value function, while ``Exploration'' designates selecting other actions randomly. 

The value function is updated as Eq.~(\ref{valuefunction})
\begin{equation}
\hat{q}\left(S_{t}, A_{t}\right)=R_{t+1}+\gamma \max _{\boldsymbol{ A }} q\left(S_{t+1}, A_{t+1}\right)\label{valuefunction}
\end{equation}
where $\hat{q}$ is the target Q function; $q$ is the estimated Q function; $R_{t+1}$ is the reward of time step $t+1$; $\gamma$ is a constant to indicate the discount rate.

No prior knowledge about the value function required at this process, instead, a more accurate value function can be fitted through experiences. The pseudo-code of choosing an action is as follows:

\begin{algorithm}
	\caption{\textit{Choosing a task by ``Exploitation'' or `` Exploration''}}\label{PC:chooseactions}
	\KwIn{State at time step $t$ $S_t$,\mbox{ } available tasks at time step $t$ $A(S_t)$,\mbox{ } and the Q-network $Q$}
	\KwOut{A selected task from all available tasks $a_t$}
	Threshold $\varepsilon$ initialization\\
	$idx \leftarrow$ random()\\
	\If{$idx \leq \varepsilon$}
	{
		$a_t \leftarrow $A task which is randomly chosen from $AL_t$\\
	}
	\Else
	{
		Calculate $Q(S_t)$\\
		$Q\_in\_list \leftarrow \emptyset$\\
		\For{$al \in \boldsymbol{ A }(S_t)$}
		{
			$Q\_in\_list \leftarrow Q\_in\_list$.append($Q(s, AL[i])$)\\
		}
		$a_t \leftarrow$ argmax $(Q\_in\_list)$\\
	}
\end{algorithm}

In the above algorithm, threshold value $\varepsilon$ is the probability of choosing an action randomly (line 3 to 4). Otherwise, the action is taken according to $Q$. The Q values of all actions at state $s_t$ are calculated by the Q-network, which is trained by Eq.~(\ref{valuefunction}). Only the actions in $\boldsymbol{ A }(S_t)$ are considered at time step $t$, so we record the Q value of all available actions $Q\_in\_list$, and then get the action with the highest Q value. see line 7 to 10.

After training, the value function (Q-network) has been finalized. In the testing process, the algorithm always select the action with the highest Q value for seeking the satisfactory solution. 

\subsection{OR algorithms in the rear stage}

Once the solution of the prior stage has been determined, there are several types of algorithms for dealing with the problem in the rear stage. Dynamic programming and a constructive heuristic algorithm are applied in this study.

\subsubsection{Constructive heuristic algorithm}

A constructive heuristic algorithm based on the density of residual tasks, named HADRT\cite{he2019scheduling}, is introduced as one of the algorithms for sequencing and timing problem. The core of HADRT are the expressions below:
\begin{equation}
f(i)=g(i)+h(i)=\sum_{i \mid r_{i}>0} 1+\frac{E T-w e_{i}}{\overline{d_{l}}}\label{HADRT1}
\end{equation}
\begin{equation}
k = \arg \max f(i)\label{HADRT2}
\end{equation}
where, $g(i)$ is the number of scheduled tasks after adding the $i$-th task; $h(i)$ equals the ratio between the remaining time length and the average duration of tasks.

Under the premise of satisfying all constraints, execution start time for each scheduled task is determined as follows:
\begin{equation}
es_k = \max (wb_k, ee_{k - 1} + ct_{k,k-1})\label{esk}
\end{equation}
where $ct_{k,k-1}$ indicates the preparation time required from the end of task $k-1$ to the start of task $k$.

The expectation of total profit in the scheme obtained by HADRT is maximized, under the following conditions:

\begin{itemize}
	\item Values of preparation time and durations of tasks are fixed or follow a determined distribution;
	\item The profit of each task is a constant;
	\item The ETW of any task is not completely covered by the ETW of another task.
\end{itemize}

The above statement has been proven by mathematical induction and reduction. The detailed process of the proof is derived in \cite{he2019scheduling}, which is not discussed here. The algorithm is organized as the sequence of the following steps: 

\textbf{Step 1}: Initialization: the latest end time of the scenario, the set of candidate tasks;

\textbf{Step 2}: Calculate $f(i)$ for each available task;

\textbf{Step 3}: Find the task $k$ by Eq.~(\ref{HADRT1}) and (\ref{HADRT2});

\textbf{Step 4}: Set the execution start time of task $k$ by Eq.~(\ref{esk});

\textbf{Step 5}: Update the latest end time of the scenario if the task $k$ is scheduled;

\textbf{Step 6}: Repeat \textbf{Step 2} to \textbf{Step 5} until all of the tasks are considered.

\subsubsection{Dynamic programming}

Dynamic programming (DP) is one of the commonly used methods for solving combinatorial optimization problems. DP guarantees the optimal solution for the timing problem in the rear stage with time complexity $O(n^2)$, under the assumptions of: 1) The sequence of tasks is fixed. 2) Time could be discretized. The above statement can be proved because the problem meets the optimality principle of dynamic programming\cite{bellman1966dynamic}.

The sequence for timing problem can be obtained by a greedy algorithm, that is, the tasks are sorted in ascending order of the end time of ETWs.

\begin{figure}[h]
	\begin{center}
		\includegraphics[width=0.48\textwidth]{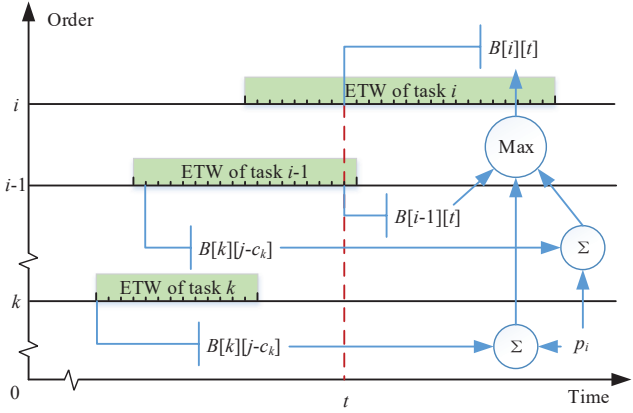}
	\end{center}
	\caption{The schematic diagram of dynamic programming in the problem.} \label{Fig:DynamicProgramming}
\end{figure}

The schematic way of applying DP to this problem is displayed in Fig.~\ref{Fig:DynamicProgramming}.

It is worth noting that this process meets the principle of optimality and without aftereffect\cite{bellman1966dynamic}. As shown in Figure~\ref{Fig:DynamicProgramming}, $B[i][t]$ is the maximum total profit that the top $i$-th tasks can get in the time $t$, which is calculated as
\begin{equation}
B[i][t]=\max _{k}\left\{B[i-1][t], B[i-k]\left[t-d_{i}-c t_{i, k}\right]+p_{i}\right\}\label{DP}
\end{equation}

Based on the above, the pseudo-code of solving the timing problem by DP is as follows:

\begin{algorithm}
	\caption{\textit{Dynamic programming}}\label{PC:dynamicprogramming}
	\KwIn{A sequence of tasks $\boldsymbol{T}$}
	\KwOut{Objective function value of the optimal solution $B[|\boldsymbol{T}|][MaxTime]$}
	Parameters initialization\\
	$MaxTime \leftarrow$ timespan\\
	\ForEach{$i \in \boldsymbol{T}$}
	{
		\For{$t=0 : MaxTime$}
		{
			\For{$k=0 : i$}
			{
				$TempB \leftarrow B[i-k][t-d_i-ct_{i,k}] + p_i$\\
				\If{satisfy all constraints \textbf{and} $TempB > B[i][t]$}
				{
					$B[i][t] \leftarrow TempB$\\
				}
				\Else
				{
					$B[i][t] \leftarrow B[i-1][t]$\\
				}
			}
		}
	}
\end{algorithm}

\subsubsection{Summary on HADRT and DP}

HADRT and DP are representative heuristic algorithm and mathematical programming algorithm in traditional OR area, respectively. Reward of time step $t$ is calculated by the result of HADRT or DP at time step $t$, therefore, the performance of HADRT and DP affects the rewards of RL. The range of rewards obtained by HADRT should be contained in the interval $\left[-\max _{i, j} p_{i}^{j}, \max _{i, j} p_{i}^{j}\right]$, in which $p_i^j$ is the profit of task $i$ on resource $j$. Correspondingly, rewards obtained by DP should be non-negative because DP guarantees the optimality of timing problem, and the range of rewards obtained by DP is reduced to $\left[0, \max _{i, j} p_{i}^{j}\right]$. 

Both the time complexity of HADRT and DP are $O(n^2)$, in which, $n$ represents the scale of the problem. Therefore, the time complexity of these two algorithms is in a low level.

We discuss the performance and consumption of these two approaches in the experiments.

\section{Experimental studies}\label{Section5}

Experiments are implemented on a personal computer with Intel(R) Core(TM) i7-8750H CPU, 16.0GB RAM and NVIDIA GeForce GTX 1060. The agile Earth observation satellite (AEOS) scheduling problem is selected as a representative example to verify the model and algorithm proposed in this paper. This problem is recognized as one of the complicated practical problems in the scheduling field\cite{mitrovic2019collection}.

\subsection{Definition of AEOS scheduling problem}

This section defines AEOS scheduling problem from inputs, objective function, and constraints.

\subsubsection{Design of resources}

Due to the periodicity of the satellite orbit, the scheduling process of the satellite can be naturally divided into orbital periods. Thereby, available orbital periods of every satellite are viewed as resources in AEOS scheduling problem. 

Refer to Eq.~(\ref{resj}), an orbital resource in AEOS scheduling problem could be described as the tuple as Eq.~(\ref{RSinAEOS}):
\begin{equation}
RS_j = (Energy_j, Storage_j)\label{RSinAEOS}
\end{equation}

A constant $Energy_j$ shows the available energy of resource $j$ and a constant $Storage_j$ indicates the available storage. In our simulation, the values of $Energy_j$ and $Storage_j$ are 150 $Ah$ and 2000 $GB$ for all resources, which are consistent with the standards of the satellite industry.

\subsubsection{Design of tasks}

A task in AEOS scheduling problem could be described by referring to Eq.~(\ref{tsij}) and Eq.~(\ref{tsi}). Under our settings, $p_i^j$ and $d_i^j$ are predetermined; $ws_i^j$ and $we_i^j$ involved in the time windows of tasks could be calculated prior to the scheduling process by considering the orbit parameters and task locations $(lat_i, lon_i)$. The discretization interval of  $ws_i^j$ and $we_i^j$ are 1 second.

The diversity of AEOS scheduling problem leads to the difficulty of finding a generally recognized benchmark that is close to actual applications. To facilitate comparisons and analyses, we create several test sets following certain distributions. Rules for generating simulated tasks are listed in Table~\ref{Tab:TaskPara}.

\begin{table}[htbp]
	\centering
	\caption{Rules for generating details of tasks}
	\begin{tabular}{p{11em}p{4.5em}p{3em}p{2.5em}p{2.5em}}
		\toprule
		Parameters & Distribution & Type & Lower bound & Upper bound\\
		\midrule
		$lat_i$ in large-area scenes & Evenly & Float & 3 & 53 \\
		$lon_i$ in large-area scenes & Evenly & Float & 73 & 133 \\
		$lat_i$ in small-area scenes & Evenly & Float & 20 & 30 \\
		$lon_i$ in small-area scenes & Evenly & Float & 108 & 114 \\
		$d_i^j$ & Constant & Integer & 5 & 5 \\
		$p_i^j$ & Evenly & Integer & 1 & 10 \\
		\bottomrule
	\end{tabular}%
	\label{Tab:TaskPara}%
\end{table}%

Two types of scenarios were designed to test the effectiveness of the proposed method in different geographical distributions of tasks. The ability to handle routine tasks could be tested in large-area scenes, which covers the whole are of China; the responsiveness of AEOS in emergencies like floods and earthquakes could be tested in small-area scenes, which covers Hunan province in China. Accurately, two small-area scenes with task sizes of 20 and 50 as well as three large-area scenes with task sizes of 100, 200, and 400 were designated, which are marked as H\_20, H\_50, C\_100, C\_200, and C\_400, respectively. 

\subsubsection{Objective and constraints}

The objective function and constraints of AEOS scheduling models are usually depended on the specific satellite platforms and application mode of the observation systems. By extensively investigating the descriptions of AEOS scheduling problems in literatures, we summarize the common objective function and some common constraints of the problem\cite{deng2017two, liu2017adaptive, wu2015coordinated, luo2017high}. 
\begin{equation}
\max \sum_{\left\{i \mid r_{i}>0\right\}} p_{i}^{r_{i}}\label{objective}
\end{equation}
subject to
\begin{equation}
\operatorname{card}\left(\left\{r_{i} \mid r_{i}>0\right\}\right) \leq 1\label{constraint1}
\end{equation}
\begin{equation}
\sum_{\left\{i \mid r_{i}=j\right\}} Storage(i, j) \leq Maxstorage(j)\label{constraint2}
\end{equation}
\begin{equation}
\sum_{\left\{i \mid r_{i}=j\right\}} Energy(i, j) \leq MaxEnergy(j)\label{constraint3}
\end{equation}
\begin{equation}
w s_{i}^{r_{i}}-e s_{i} \leq 0\label{constraint4}
\end{equation}
\begin{equation}
e s_{i}+d_{i}^{r_{i}}-w e_{i}^{r_{i}} \leq 0\label{constraint5}
\end{equation}
\begin{equation}
\left(e s_{i_{0}}-e s_{i_{1}}\right)\left(e s_{i_{0}}-e e_{i_{1}}\right)>0 \quad \forall i_{0} \neq i_{1}, r_{i_{0}}=r_{i_{1}}\label{constraint6}
\end{equation}
\begin{equation}
\left(e e_{i_{0}}-e s_{i_{1}}\right)\left(e e_{i_{0}}-e e_{i_{1}}\right)>0 \quad \forall i_{0} \neq i_{1}, r_{i_{0}}=r_{i_{1}}\label{constraint7}
\end{equation}
\begin{equation}
\left|e s_{i_{0}}-e e_{i_{1}}\right| \geq MST\left(i_{0}, i_{1}\right) \quad \forall i_{0} \neq i_{1}, r_{i_{0}}=r_{i_{1}}\label{constraint8}
\end{equation}
\begin{equation}
i = 1,2,\ldots,n\label{definitiondomain1}
\end{equation}
\begin{equation}
j = 1,2,\ldots,m\label{definitiondomain2}
\end{equation}

In the above model, the objective function is to maximize the sum of the profit of all scheduled tasks, which as~(\ref{objective}). Inequality~(\ref{constraint1}) to~(\ref{constraint3}) affiliate to constraints in the prior stage, while~(\ref{constraint4}) to (\ref{constraint8}) belong to that of the rear stage.

Constraints~(\ref{constraint1}) refers that a task is scheduled at most once;~(\ref{constraint2}) and~(\ref{constraint3}) shows the capacity constraints of the AEOSs with respect to energy and storage.~(\ref{constraint4}) and~(\ref{constraint5}) indicate that all tasks should be executed in one of their ETWs.~(\ref{constraint6}) to~(\ref{constraint8}) expresses the constraints between two tasks, i.e., no overlap in the execution time of any two scheduled tasks is allowed. Meanwhile, the interval between the two tasks must be greater than the corresponding slew time, to guarantee the shortest time required by the satellite to adjust the attitude from a task to the next one. 

The configurations of slew time, storage consumption, and electricity consumption are derived from a practical example in China, which named AS-01. The necessary functions of AS-01 are detailed in~\cite{liu2017adaptive, he2016research}.

\subsection{Performance of applying Reinforcement Learning-based algorithms to AEOS scheduling}

The feasibilities of the proposed algorithms are tested in this section. In addition to DQN\_CH and DQN\_DP that are introduced in Section~\ref{Section4}, the experiment introduces two other approaches for comparison of results, that is:

\textbf{DQN\_CH/C}: DQN for assignment problem and sequencing problem, while HADRT for timing problem;

\textbf{DQN\_DP/C}: DQN for assignment problem and sequencing problem, while DP for timing problem;

\subsubsection{Convergence analysis}

\begin{figure}[htbp]
	\centering
	\subfigure[Results on H\_20]{
		\begin{minipage}[t]{0.5\linewidth}
			\centering
			\includegraphics[width=\linewidth]{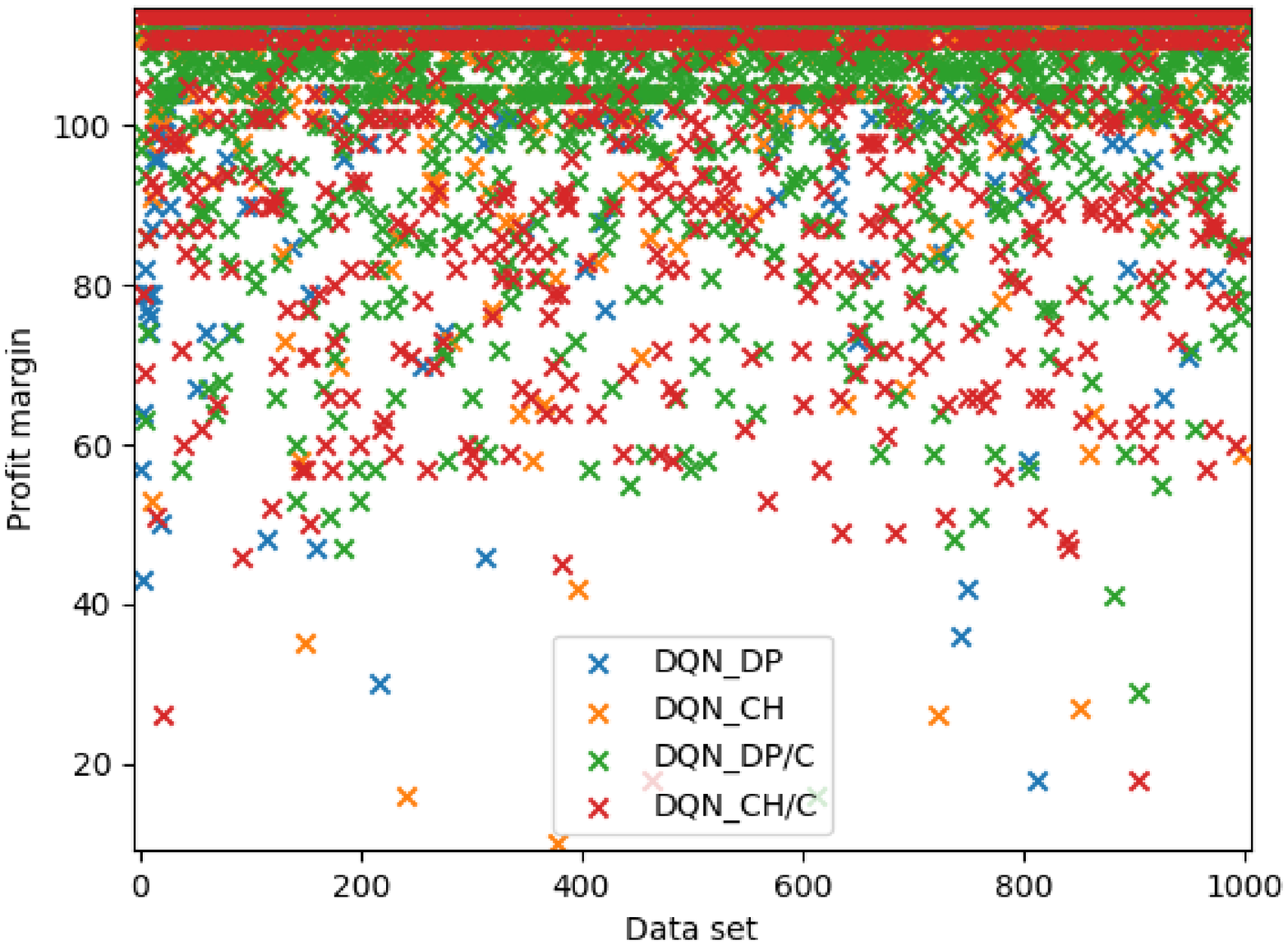}
		\end{minipage}%
	}%
	\subfigure[Results on H\_50]{
		\begin{minipage}[t]{0.5\linewidth}
			\centering
			\includegraphics[width=\linewidth]{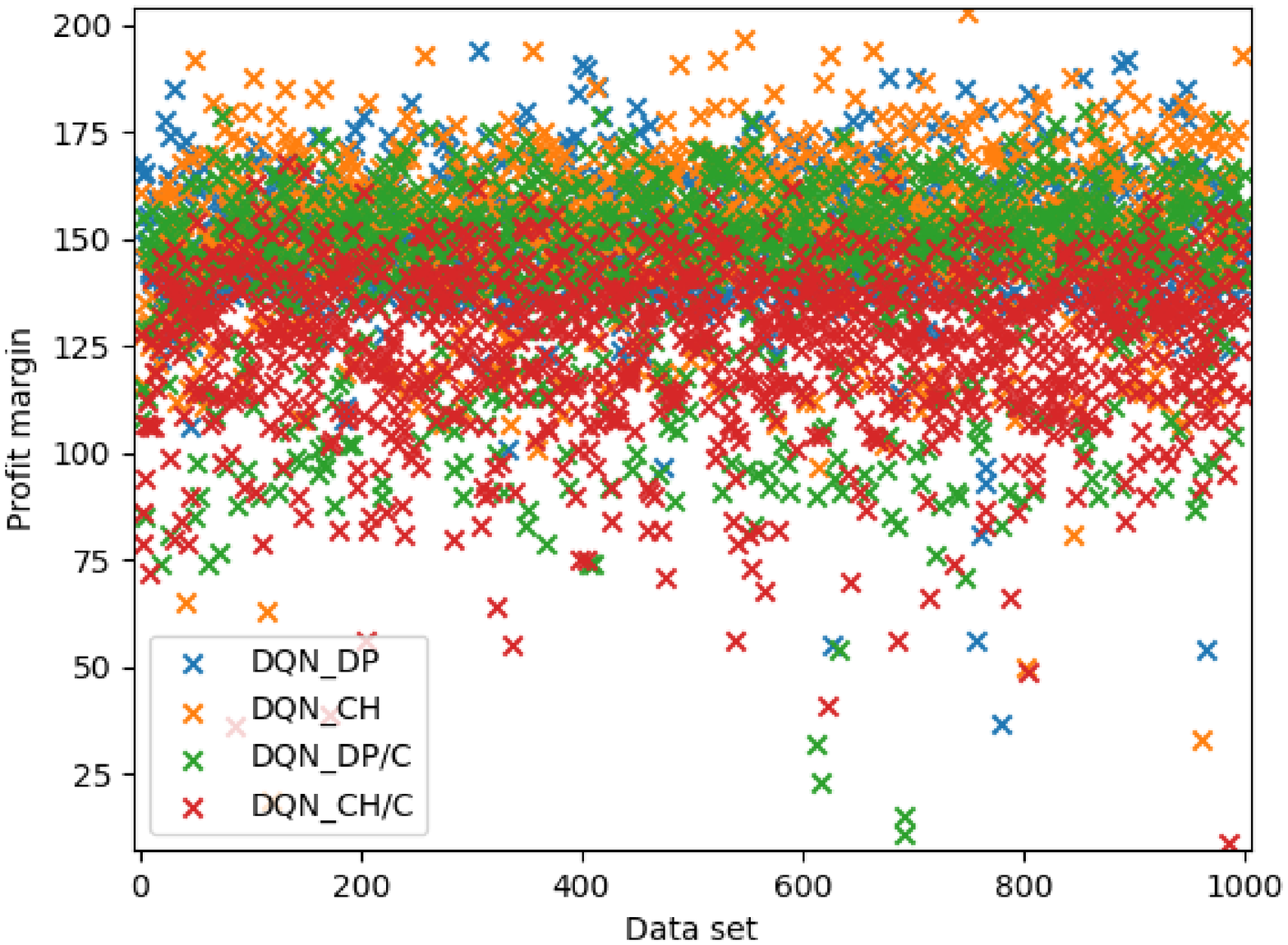}
		\end{minipage}%
	}%

	\subfigure[Results on H\_100]{
		\begin{minipage}[t]{0.5\linewidth}
			\centering
			\includegraphics[width=\linewidth]{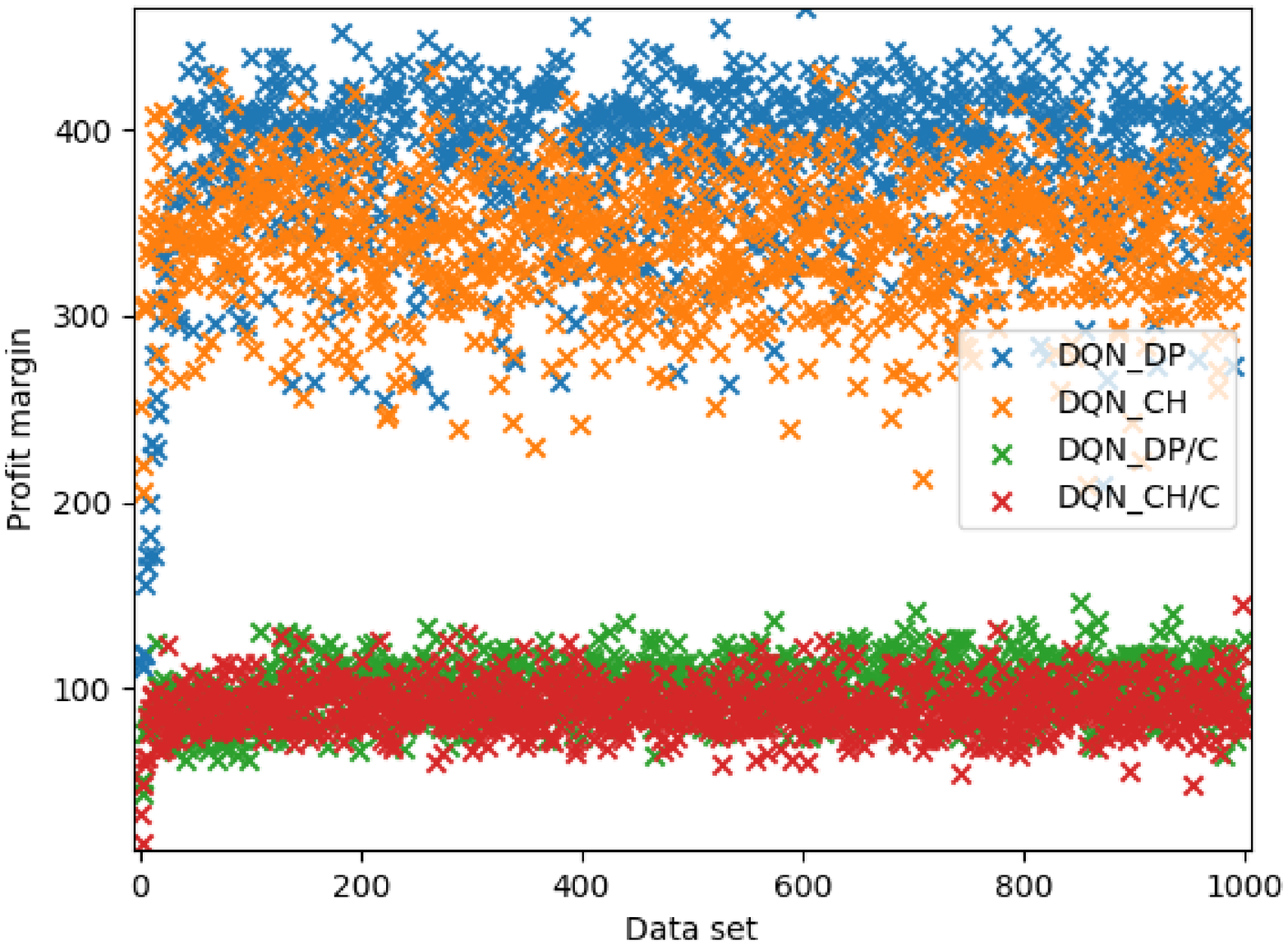}
		\end{minipage}%
	}%
	\subfigure[Results on H\_200]{
		\begin{minipage}[t]{0.5\linewidth}
			\centering
			\includegraphics[width=\linewidth]{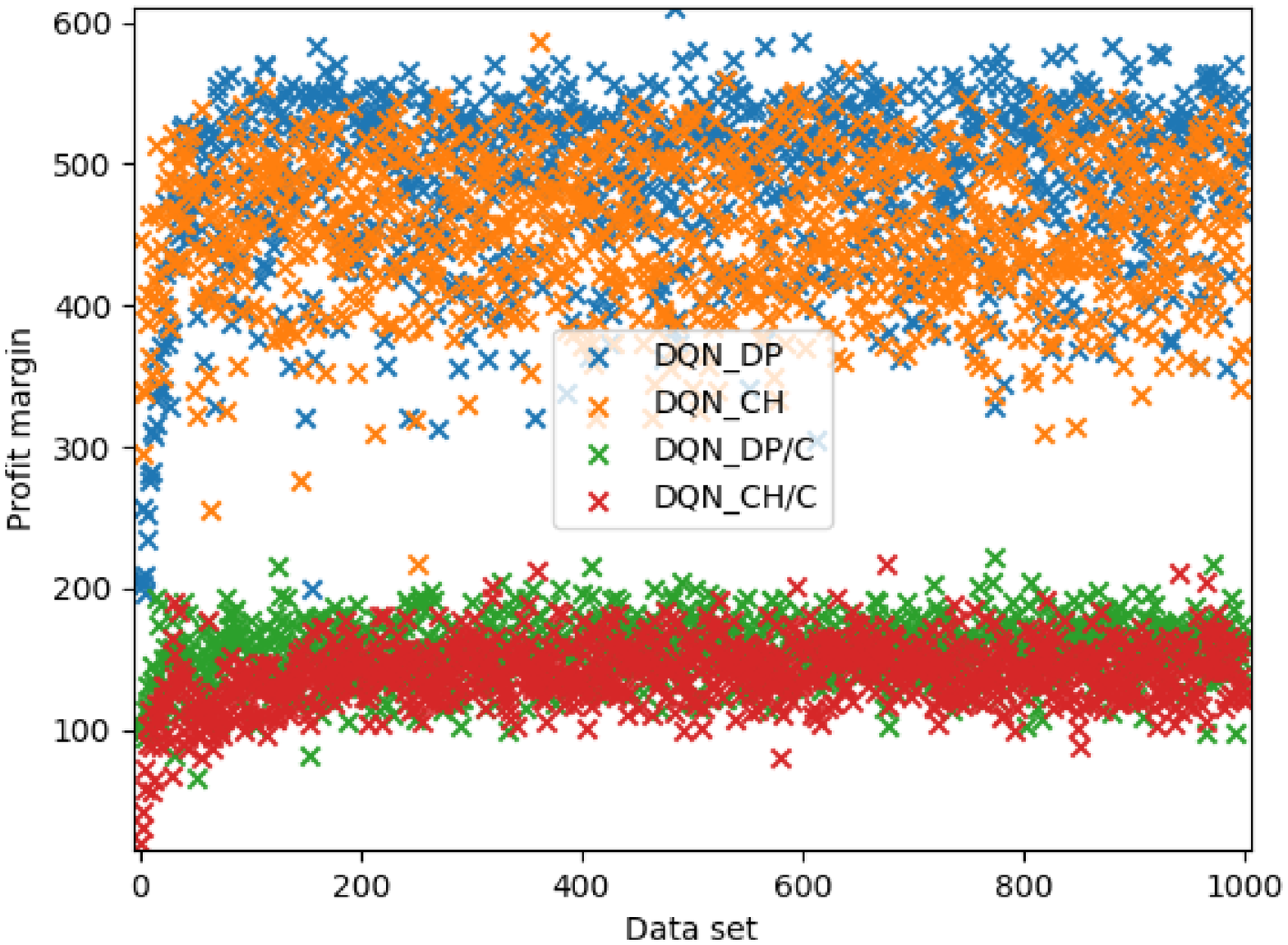}
		\end{minipage}
	}%

	\subfigure[Results on H\_400]{
		\begin{minipage}[t]{0.5\linewidth}
			\centering
			\includegraphics[width=\linewidth]{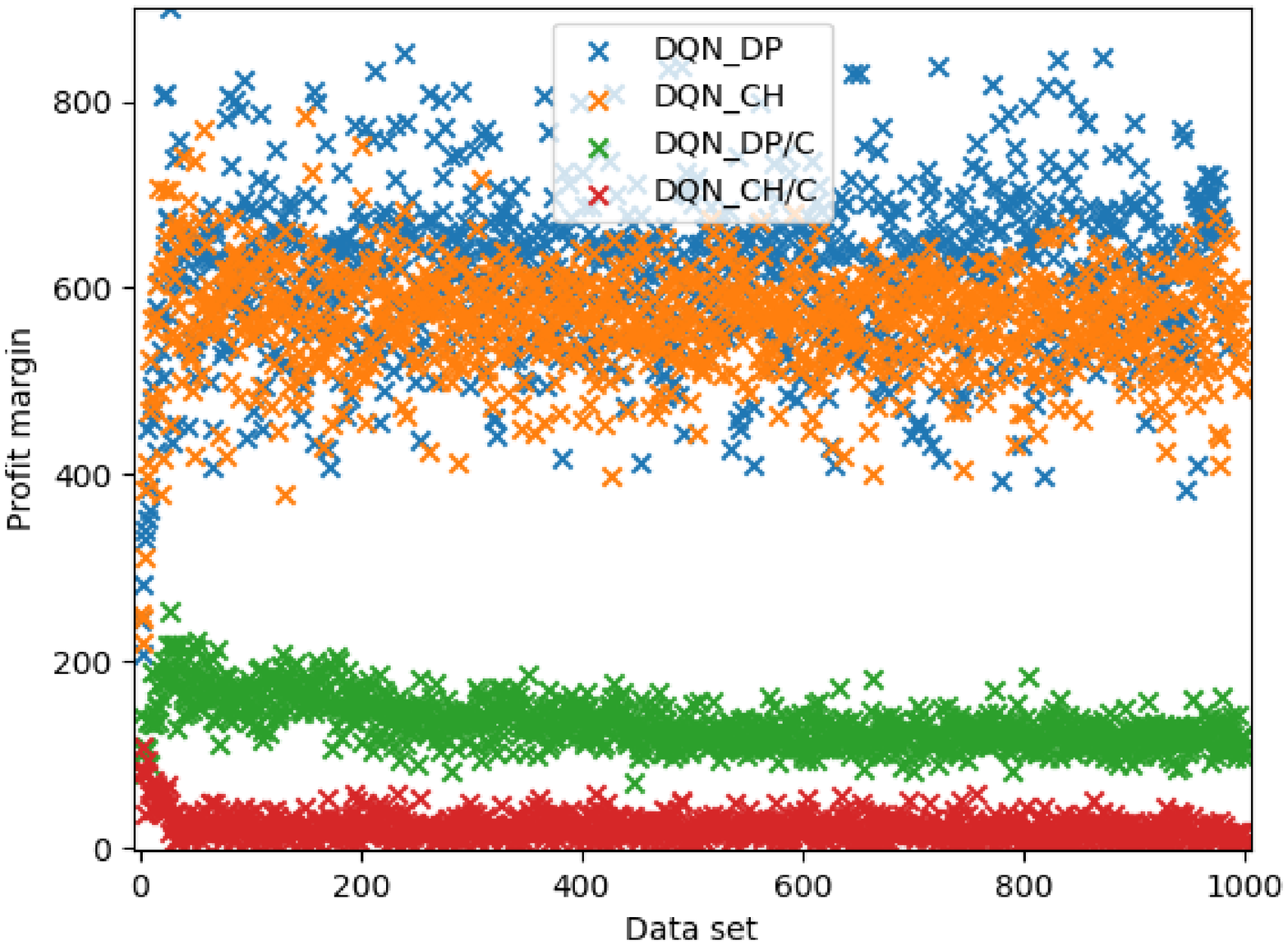}
		\end{minipage}
	}%
	
	\centering
	\caption{Total profit of proposed algorithms under 1000 iterations in single instance.}\label{Fig:Convergence}
\end{figure}

1000 iterations on the same instance are used to verify the convergence of the proposed methods, as shown in Fig.~\ref{Fig:Convergence}. Through dozens of iterations, the total profit of all algorithms is stable overall, with acceptable fluctuations. The fluctuation is mainly caused by the existence of ``Exploration'' in the training process of DQN.

As the number of tasks increases, the differences between these four algorithms are apparent: The average total profit of DQN\_CH and DQN\_DP in all scenarios is higher than DQN\_DP/C and DQN\_CH/C. Furthermore, DQN\_DP/C and DQN\_CH/C will converge to a low level on C\_100, C\_200, and C\_400. These two algorithms are easily trapped in a local optimum solution, and they did not show the trend to escape the local optimum in the first 1000 iterations. The results show the sequencing problem is a high-complexity problem for DQN. Thus, these two algorithms are hard to achieve satisfactory results through small-scale training. The total profit of DQN\_DP averages around 10\% higher than that of DQN\_CH, but DQN\_DP requires more computing resources and time during the training process: it can only complete 949 iterations in C\_400 because of memory overflow.

\subsubsection{Generalization analysis}

\begin{figure}[htbp]
	\centering
	\subfigure[Results on H\_20]{
		\begin{minipage}[t]{0.5\linewidth}
			\centering
			\includegraphics[width=\linewidth]{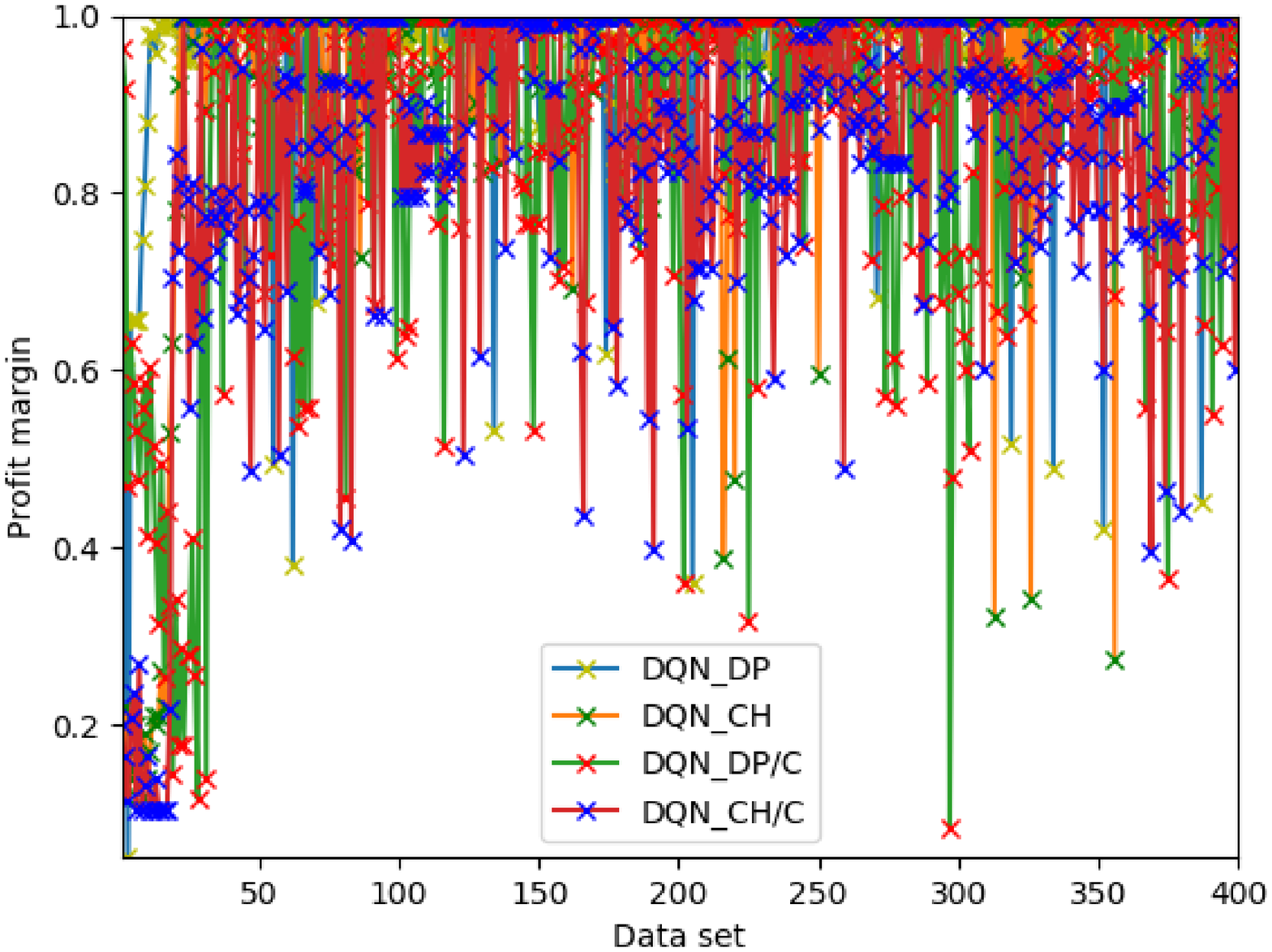}
		\end{minipage}%
	}%
	\subfigure[Results on H\_50]{
		\begin{minipage}[t]{0.5\linewidth}
			\centering
			\includegraphics[width=\linewidth]{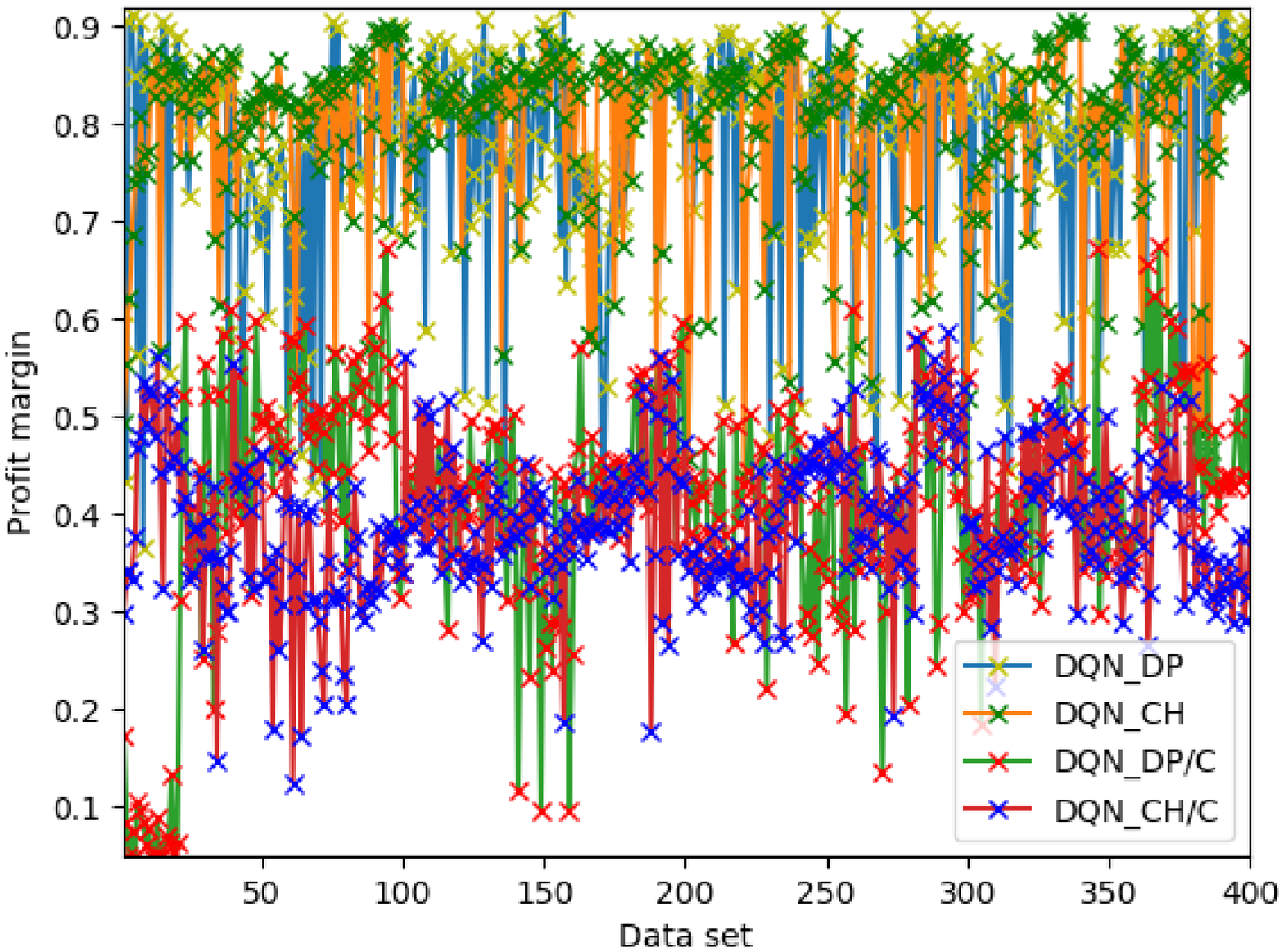}
		\end{minipage}%
	}%
	
	\subfigure[Results on H\_100]{
		\begin{minipage}[t]{0.5\linewidth}
			\centering
			\includegraphics[width=\linewidth]{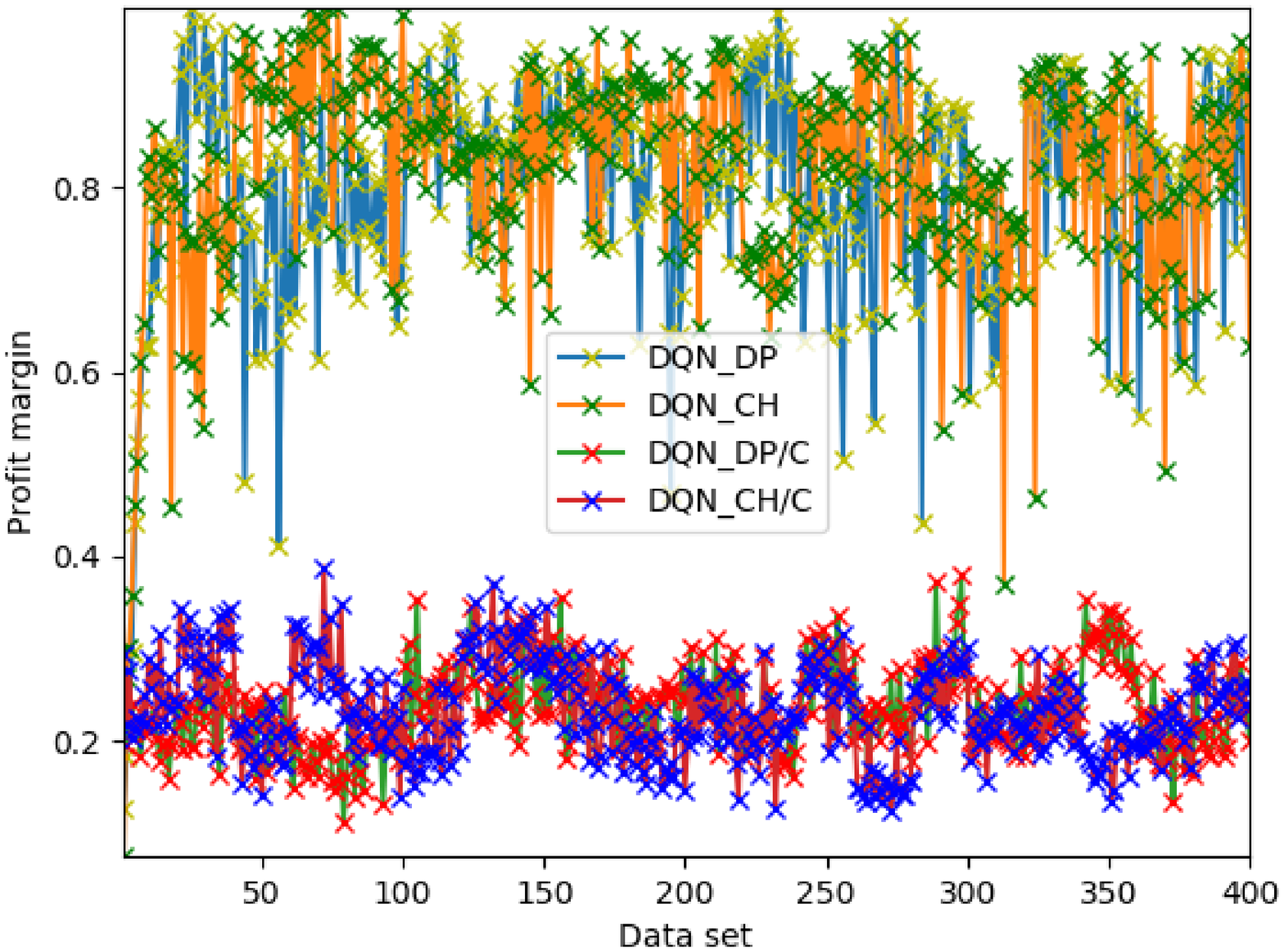}
		\end{minipage}%
	}%
	\subfigure[Results on H\_200]{
		\begin{minipage}[t]{0.5\linewidth}
			\centering
			\includegraphics[width=\linewidth]{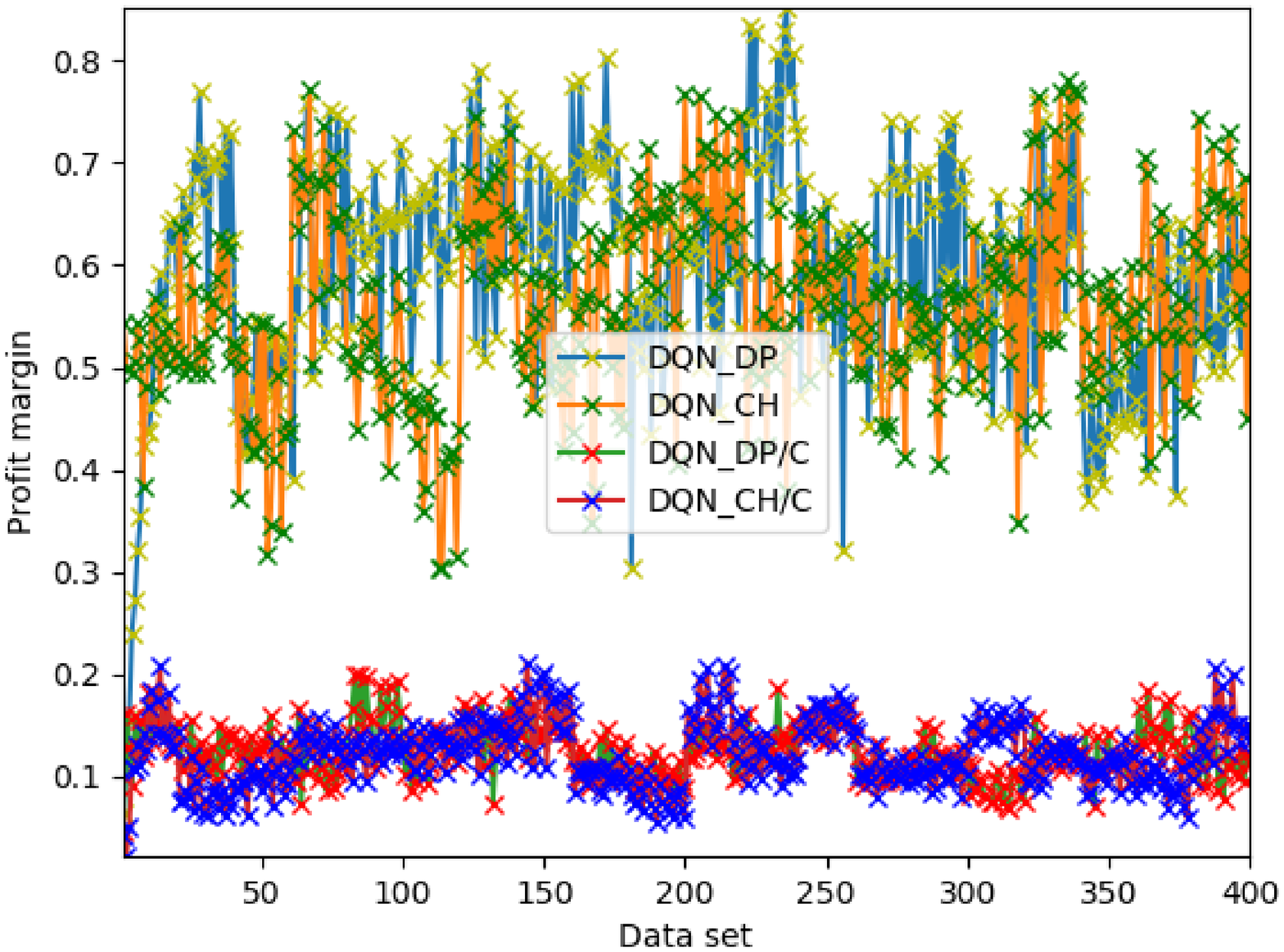}
		\end{minipage}
	}%

	\subfigure[Results on H\_400]{
		\begin{minipage}[t]{0.5\linewidth}
			\centering
			\includegraphics[width=\linewidth]{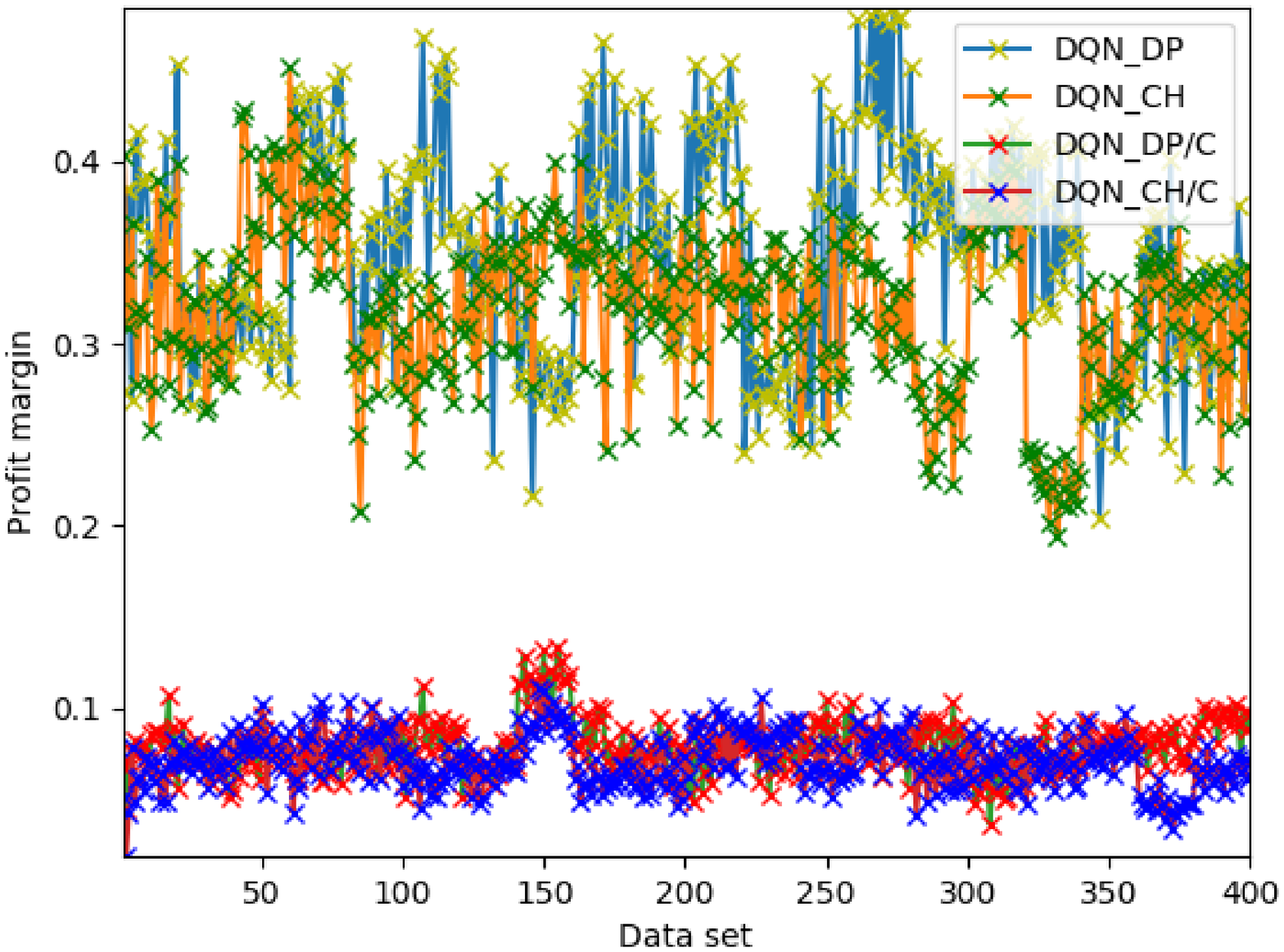}
		\end{minipage}
	}%
	
	\centering
	\caption{Total profit of proposed algorithms under 20 iterations in multiple instances.}\label{Fig:Generalization1}
\end{figure}

20 random instances are used to test the generalization of the proposed methods. Depending on the different scales, training processes of different instances take tens of minutes to several hours, including the time for preprocessing and status updates. Since the profit of each task in a task set is random, analyzing the trend of profit margin is more reasonable. The profit margin is the ratio of the total profit of scheduled tasks to that of all candidate tasks. As can be seen in Fig.~\ref{Fig:Generalization1}, although only 20 iterations are performed on each data set, the polyline of each algorithm fluctuates across a certain level, indicating the value functions trained by all four algorithms are applicable in different scenarios, that is, these four algorithms are robust in generalization. However, the fluctuations in Fig.~\ref{Fig:Generalization1} are larger than the fluctuation range in Fig.~\ref{Fig:Convergence}. This is because: 1) the optimal solutions vary in each instance; 2) the value functions need to be continuously modified in different instances.

\begin{figure}[h]
	\begin{center}
		\includegraphics[width=0.48\textwidth]{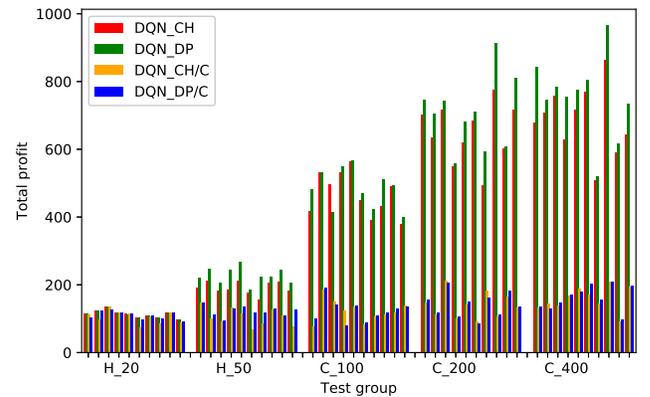}
	\end{center}
	\caption{Total profit of proposed algorithms on test data.} \label{Fig:Generalization2}
\end{figure}

The stability of the results in distinct instances implies the generalization. Algorithms trained from 20 instances are applied to unknown scenes, and results are shown in Fig.~\ref{Fig:Generalization2}. In 50 instances with different features, the test results of the four algorithms are consistent with the rules during training. The results of DQN\_CH and DQN\_DP are much better than those of the other two. In all 50 test sets, DQN\_CH are superior to DQN\_DP on just two instances; except for the instances in H\_20, the result of DQN\_CH and DQN\_DP equally on only one instance.

\subsection{Comparison with non-learning scheduling algorithms}

We compared DQN\_CH and DQN\_DP to several advanced algorithms that have published in recent papers. These algorithms include a heuristic represented by a heuristic algorithm based on the density of residual tasks\cite{he2019scheduling}, a meta-heuristic represented by the adaptive large neighborhood search algorithm\cite{liu2017adaptive}, and a mathematical programming method represented by the branch and bound algorithm\cite{chu2017branch}, which are denoted as HADRT, ALNS, and B\&B, respectively. 

\subsubsection{Total profit}

\begin{figure}[h]
	\begin{center}
		\includegraphics[width=0.48\textwidth]{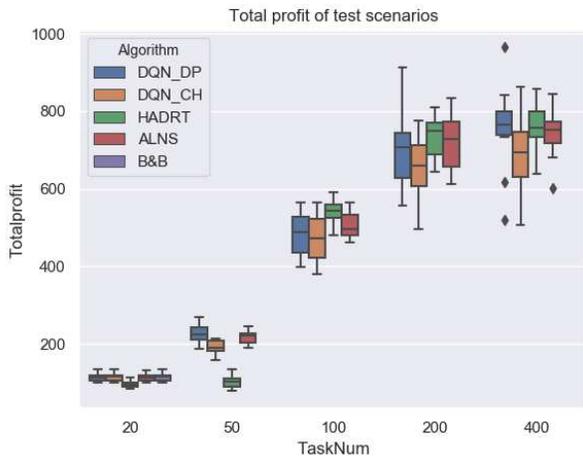}
	\end{center}
	\caption{Comparison of the RL-based optimization algorithms with advanced scheduling algorithms on the test data.} \label{Fig:TotalProfitComparison}
\end{figure}

In this problem, total profit is the value of the objective function, which shows the optimization capability of an algorithm. Thus, it is a vital indicator for evaluating algorithms. We ran concerned algorithms in all test sets and recorded results whose running time less than 3600 seconds. All results are summarized in Fig.~\ref{Fig:TotalProfitComparison}. The box plot shows: 1) B\&B could obtain the optimal solution in small scale problems (in test set H\_20), but cannot obtain the result within 3600 seconds in other test sets; 2) HADRT could reach good results for most instances, but in some cases, the results are not satisfactory; 3) ALNS, DQN\_CH, and DQN\_DP can reliably achieve high-quality results, especially for large test sets. In general, DQN\_DP performs the best among these five comparative algorithms.

\subsubsection{Running time}

\begin{table*}
	\centering
	\caption{The average and standard deviation of running time on different data sets.}
	\begin{tabular}{ccccccccccc}
		\toprule
		\multirow{2}[4]{*}{Scenes} & \multicolumn{5}{c}{MEAN} & \multicolumn{5}{c}{STDEV} \\
		\cmidrule{2-11}          & DQN\_CH & DQN\_DP & HADRT & ALNS  & B\&B  & \multicolumn{1}{p{4.055em}}{DQN\_CH} & \multicolumn{1}{p{4.055em}}{DQN\_DP} & \multicolumn{1}{p{4.055em}}{HADRT} & \multicolumn{1}{p{4.055em}}{ALNS} & B\&B \\
		\midrule
		H\_20 & 0.079 & 1.824 & 0.038 & 7.777 & 1.553 & 0.004 & 0.038 & 0.013 & 1.146 & \multicolumn{1}{c}{1.234} \\
		H\_50 & 0.153 & 5.539 & 0.126 & 20.287 & $>$3600 & 0.028 & 0.38  & 0.011 & 11.653 & $-$ \\
		C\_100 & 0.399 & 9.367 & 0.152 & 38.691 & $>$3600 & 0.024 & 0.752 & 0.017 & 26.916 & $-$ \\
		C\_200 & 1.003 & 24.655 & 0.442 & 316.83 & $>$3600 & 0.12  & 1.44  & 0.034 & 75.318 & $-$ \\
		C\_400 & 1.809 & 24.426 & 1.246 & 2057.9 & $>$3600 & 0.234 & 2.031 & 0.098 & 222.41 & $-$ \\
		\bottomrule
	\end{tabular}%
	\label{Tab:CPUtime}%
\end{table*}%

For practical problems, the running time of the program is another important metric to evaluate the efficiency of the algorithm. According to the data in Table~\ref{Tab:CPUtime}, HADRT, DQN\_CH, and DQN\_DP could get results in a short time, while the running time of ALNS and B\&B grow dramatically with the increase of the problem size. The computation time of DQN\_CH is approximately twice that of HADRT. Although the running time of DQN\_DP is much larger than that of DQN\_CH and HADRT, they increase linearly with the scale of the problem. By analyzing the standard deviation of running time, it can be found that under the same test group, the computation time of ALNS and B\&B is greatly affected by the characteristics of inputs. By contrast, HADRT, DQN\_CH, and DQN\_DP run with a stable time.

Overall, B\&B only works well on small-scale instances. HADRT could find a feasible solution in a short time, but the quality of the solution could not be guaranteed. For ALNS, the risk of falling into a local optimum can be reduced through the random search mechanism, but it usually takes too long to search a promising solution, which may be unacceptable in many practical problems. RL-based optimization algorithms could be a potential alternative to obtain satisfactory solutions with acceptable time.

\section{Conclusions}\label{Section6}

We have summarized the essential characteristics of a type of complex scheduling problems, and then propose a two-stage scheduling framework to solve them. Based on the framework, we have built two RL-based optimization algorithms. These algorithms are designed by combining DQN with different OR algorithms (i.e., HADRT or DP). Experimental results demonstrate the effectiveness and advantage of RL-based optimization algorithms.

This is a preliminary attempt about combining the machine learning and operations research to solve a practical scheduling problem. A two-stage decision model is devised to describe general scheduling problem. Decision variables are divided into two parts and their values are determined in two stages. The assignment problem can be seen as the prior stage and solved by RL, while the OR algorithms can be used to solve the sub-problems in the the rear stage. It is more likely to fall into a local optimum when solving the sequencing problem in the prior stage, and it is difficult to jump out of the local optimum in small-scale training. The impact of different OR algorithms used in the rear stage is analyzed. The reward obtained by DP is closer to the optimal solution of the problem in the rear stage than that of HADRT, thereby DQN\_DP could perform better. However, DQN\_DP consumes relatively more computing resources both in training and testing. 

Although deep Q-learning is one of the basic methods in RL, this work shows the great potential in solving complex combinatorial optimization problems by the RL-based optimization algorithms. To further enrich the conclusions, we will try more RL methods and structures of the value function, and delve into the inherent rules of the integrating learning algorithms with traditional optimization algorithms. Furthermore, we will continue to test the performance of RL-based optimization algorithms on other types of scheduling problems.

\section{Acknowledgments}

This research is supported by the Natural Science Fund for Distinguished Young Scholars of Hunan Province under Grant 2019JJ20026 and the National Natural Science Foundation of China under Grant No.71701204.


\bibliographystyle{IEEEtran}
\bibliography{mybib}

%
\begin{IEEEbiography}[{\includegraphics[width=1.0in,height=1.5in, clip, keepaspectratio]{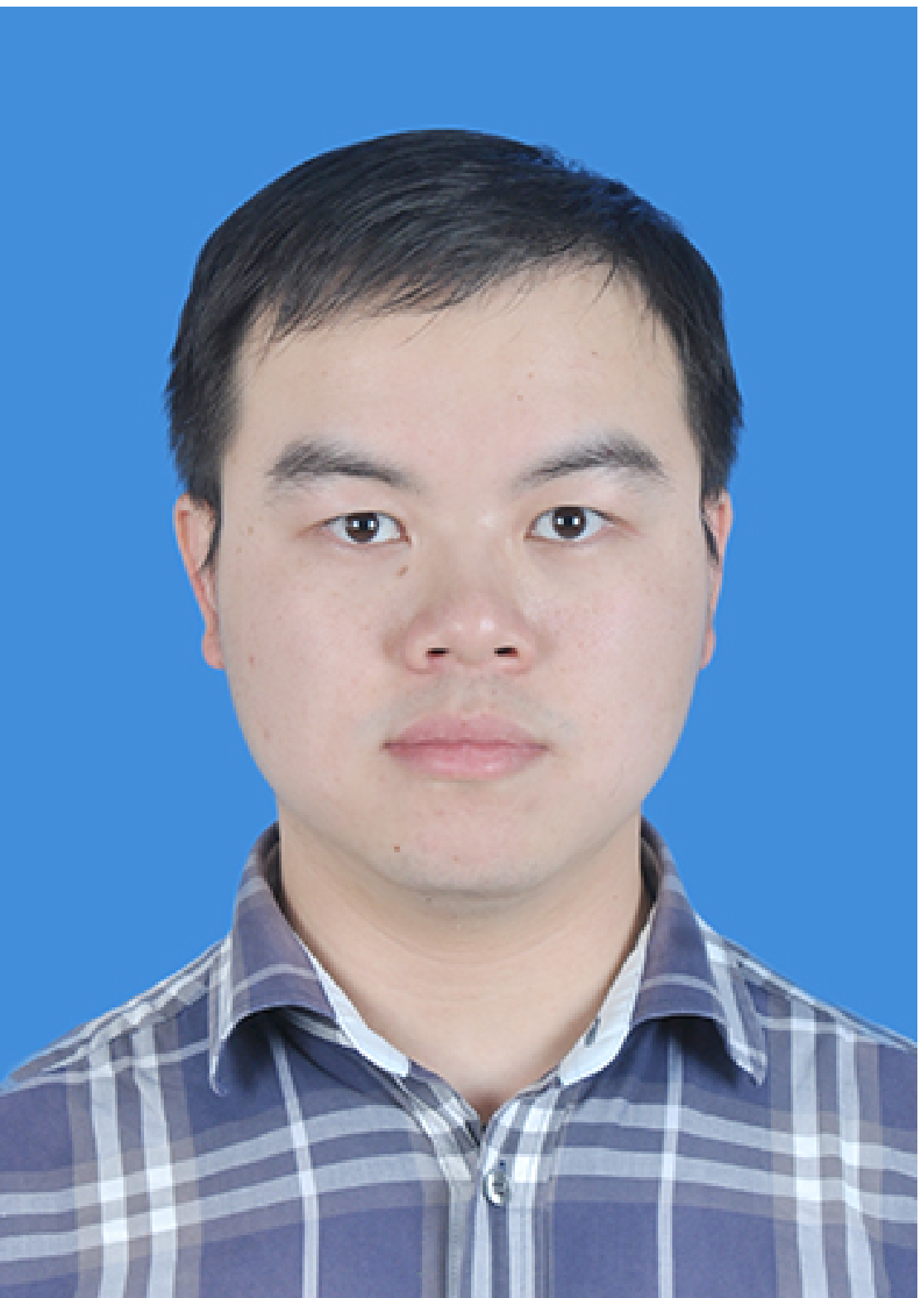}}] {Yongming He} received the B.S. degree in Logistics Engineering from Chang'an University, China, in 2014. He is now pursuing his Ph.D degree in Management Science and Engineering at the College of Systems Engineering, National University of Defense Technology, China. He is a visiting Ph.D. student at University of Alberta, Edmonton, AB, Canada, from Nov. 2018 to Nov. 2019. His research interests include operations research, artificial intelligence, intelligent decision, scheduling and planning.
\end{IEEEbiography}

\begin{IEEEbiography}[{\includegraphics[width=1.0in,height=1.5in, clip, keepaspectratio]{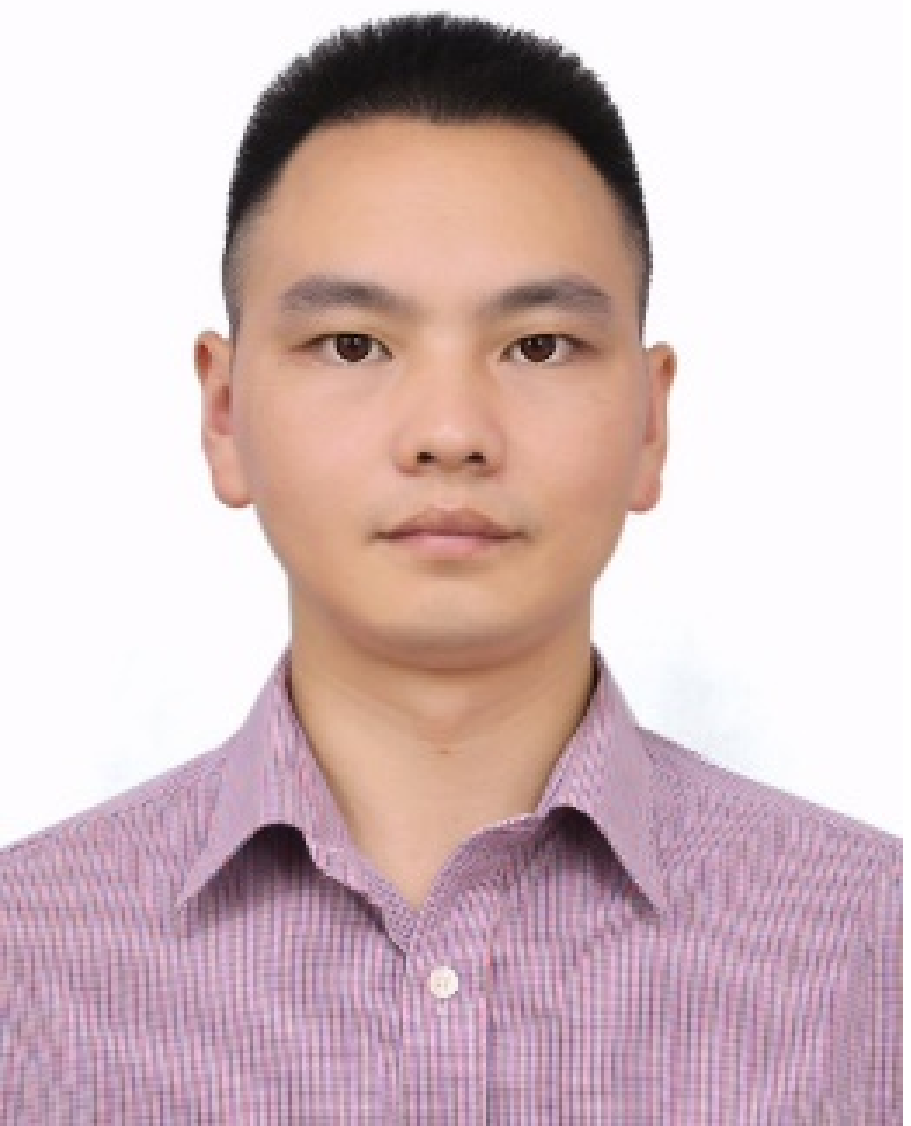}}] {Guohua Wu} received the B.S. degree in Information Systems and Ph.D degree in Operations Research from National University of Defense Technology, China, in 2008 and 2014, respectively. During 2012 and 2014, he was a visiting Ph.D student at University of Alberta, Edmonton, Canada. He is currently a Professor at the School of Traffic and Transportation Engineering, Central South University, Changsha, China.
	
His current research interests include Planning and Scheduling, Computational Intelligence and Machine Learning. He has authored more than 60 referred papers including those published in \textit{IEEE TCYB}, \textit{IEEE TSMCA}, \textit{Information Sciences} and \textit{Computers \& Operations Research}. He serves as an Associate Editor of \textit{Swarm and Evolutionary Computation Journal}, an editorial board member of \textit{International Journal of Bio-Inspired Computation}, and a Guest Editor of \textit{Information Sciences} and \textit{Memetic Computing}. He is a regular reviewer of more than 20 journals including \textit{IEEE TEVC}, \textit{IEEE TCYB}, \textit{Information Sciences} and \textit{Applied Soft Computing}.
\end{IEEEbiography}

\begin{IEEEbiography}[{\includegraphics[width=1.0in,height=1.5in, clip, keepaspectratio]{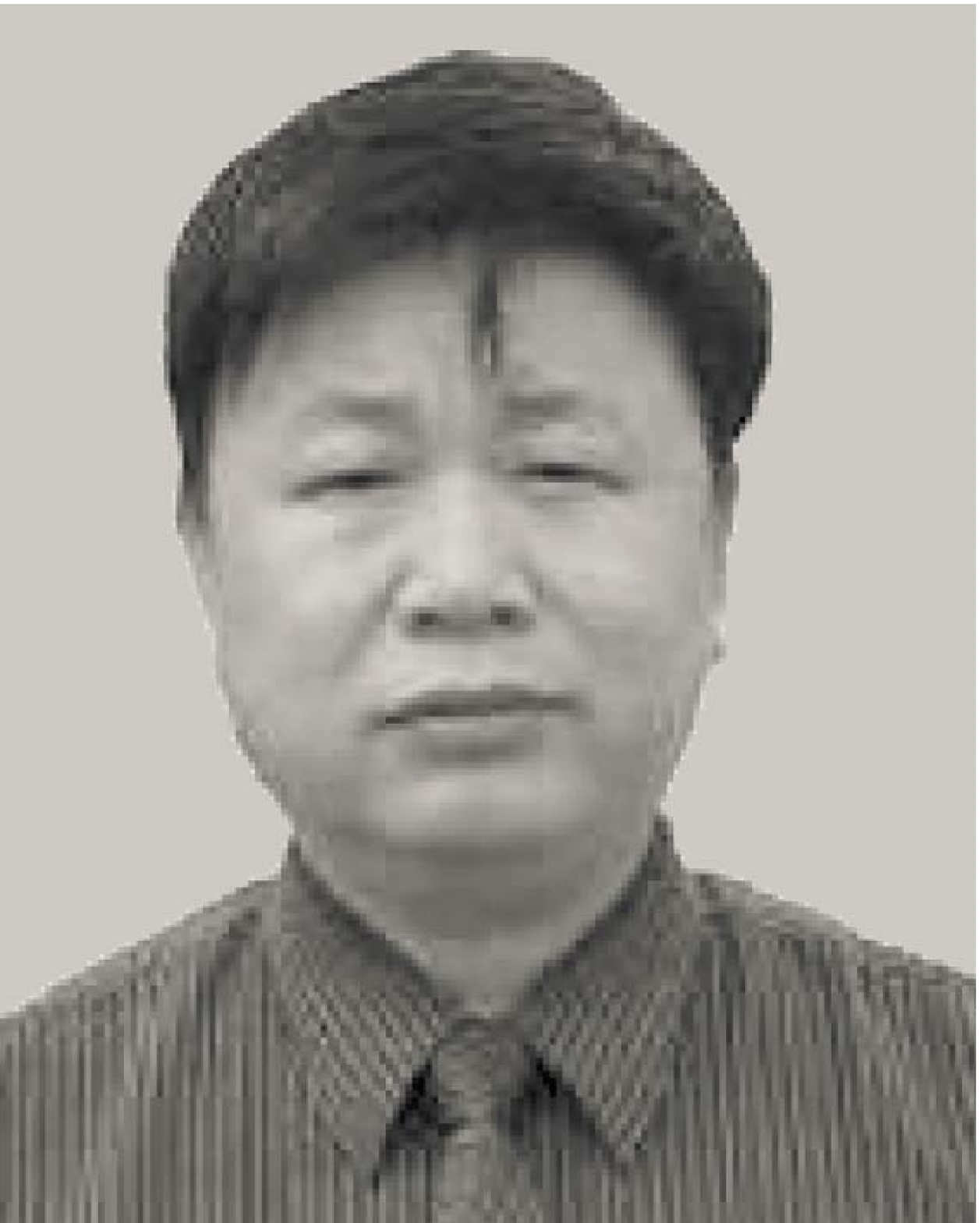}}] {Yingwu Chen}received the B.S. degree in automation, the M.S. degree in system engineering, and the Ph.D. degree in engineering from the National University of Defense Technology (NUDT), Changsha, China, in 1984, 1987, and 1994, respectively.

He was a Lecturer from 1989 to 1994, and an Associate Professor from 1994 to 1999 at NUDT. From 1999 to 2017, he was a Professor and the Director of the Department of Management Science and Engineering, College of Information Systems and Management, NUDT. Now, he has been a Distinguished Professor of the College of Systems Engineering, NUDT, where he focuses on management theory and its applications. He has authored more than 70 research publications. His current research interests include assistant decision-making systems for planning, decision-making systems for project evaluation, management decisions, and artificial intelligence.

Dr. Chen is the Editor of the Principle of System Engineering (Press of National University of Defense Technology), and Technology of Quantificational
Analysis (China Renmin University Press).
\end{IEEEbiography}

\begin{IEEEbiography}[{\includegraphics[width=1.0in,height=1.5in, clip, keepaspectratio]{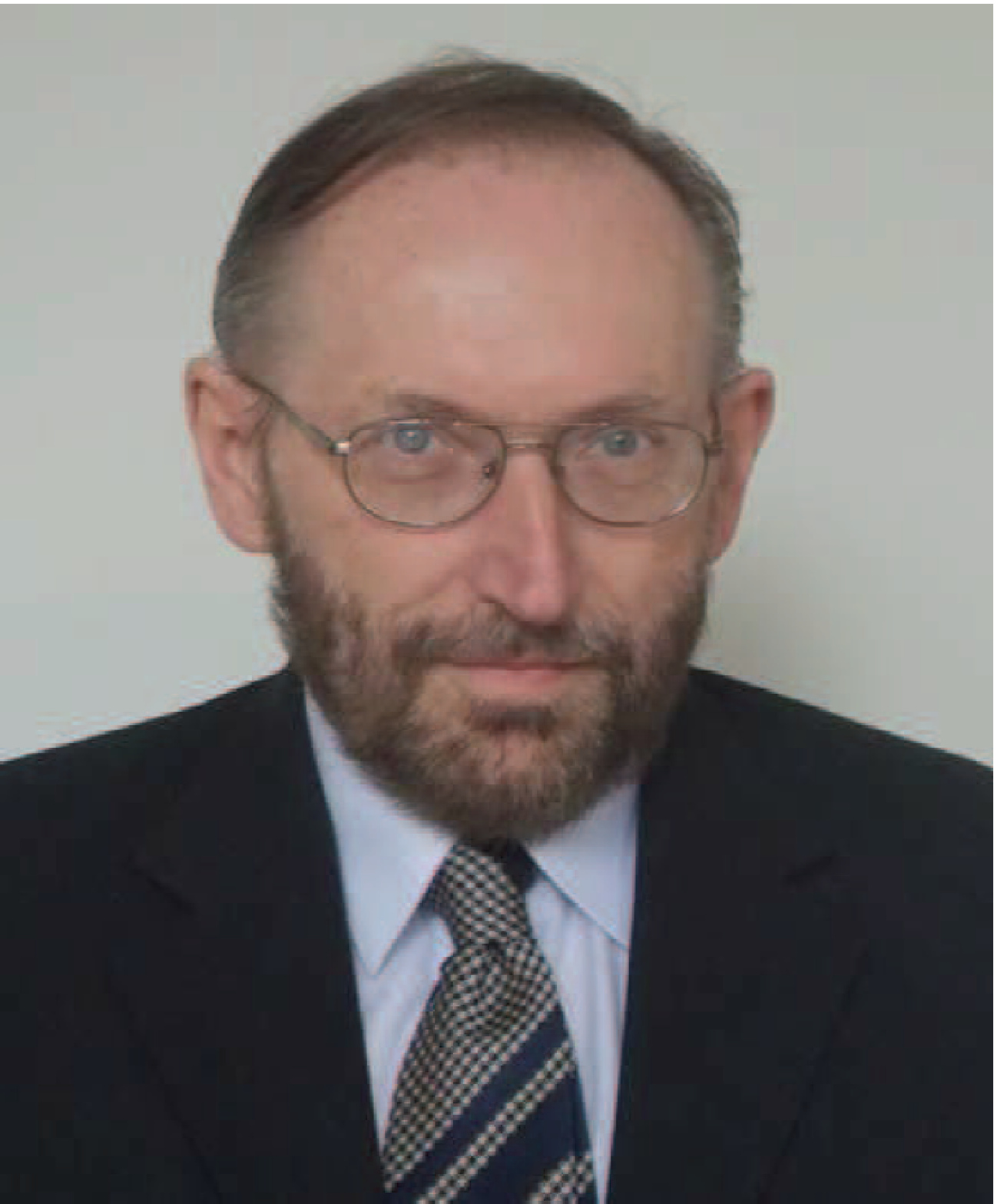}}] {Witold Pedrycz} (IEEE Fellow, 1998) is Professor and Canada Research Chair (CRC) in Computational Intelligence in the Department of Electrical and Computer Engineering, University of Alberta, Edmonton, Canada. He is also with the Systems Research Institute of the Polish Academy of Sciences, Warsaw, Poland. In 2009 Dr. Pedrycz was elected a foreign member of the Polish Academy of Sciences. In 2012 he was elected a Fellow of the Royal Society of Canada. In 2007 he received a prestigious Norbert Wiener award from the IEEE Systems, Man, and Cybernetics Society. He is a recipient of the IEEE Canada Computer Engineering Medal, a Cajastur Prize for Soft Computing from the European Centre for Soft Computing, a Killam Prize, and a Fuzzy Pioneer Award from the IEEE Computational Intelligence Society. 

His main research directions involve Computational Intelligence, fuzzy modeling and Granular Computing, knowledge discovery and data science, pattern recognition, data science, knowledge-based neural networks, and control engineering. He has published numerous papers in these areas; the current h-index is 107 (Google Scholar). He is also an author of 18 research monographs and edited volumes covering various aspects of Computational Intelligence, data mining, and Software Engineering. 

Dr. Pedrycz is vigorously involved in editorial activities. He is an Editor-in-Chief of Information Sciences, Editor-in-Chief of WIREs Data Mining and Knowledge Discovery (Wiley), and Co-editor-in-Chief of Int. J. of Granular Computing (Springer) and J. of Data Information and Management (Springer).  He serves on an Advisory Board of IEEE Transactions on Fuzzy Systems and is a member of a number of editorial boards of international journals. 
\end{IEEEbiography}

\end{document}